\documentclass{article}

\usepackage{microtype}
\usepackage{graphicx}
\usepackage{subfigure}
\usepackage{booktabs} 

\usepackage{hyperref}

 \usepackage[accepted]{mlsys2025}

\usepackage{booktabs}       
\usepackage{amsfonts}       
\usepackage{bm}
\usepackage{enumitem}
\usepackage{xcolor}

\usepackage{subfigure}
\usepackage{wrapfig}

\usepackage{amsmath}
\usepackage{amssymb}
\usepackage{mathtools}
\usepackage{amsthm}
\DeclareMathAlphabet \mathbfcal{OMS}{cmsy}{b}{n}

\newcommand{\ten}[1]{\mathbfcal{#1}} 
\newcommand{\mat}[1]{\mathbf{#1}}

\usepackage{subfiles}
\usepackage{float}

\mlsystitlerunning{Scalable Back-Propagation-Free Training of Optical Physics-Informed Neural Networks}

\begin{document}

\twocolumn[
\mlsystitle{Scalable Back-Propagation-Free Training of Optical Physics-Informed Neural Networks}



\mlsyssetsymbol{equal}{*}

\begin{mlsysauthorlist}
\mlsysauthor{Yequan Zhao}{equal,ucsb}
\mlsysauthor{Xinling Yu}{equal,ucsb}
\mlsysauthor{Xian Xiao}{hpe}
\mlsysauthor{Zhixiong Chen}{ucsb}
\mlsysauthor{Ziyue Liu}{ucsb}
\mlsysauthor{Geza Kurczveil}{hpe}
\mlsysauthor{Raymond G. Beausoleil}{hpe}
\mlsysauthor{Sijia Liu}{msu}
\mlsysauthor{Zheng Zhang}{ucsb}
\end{mlsysauthorlist}

\mlsysaffiliation{ucsb}{Department of Electrical and Computer Engineering, University of California, Santa Barbara}
\mlsysaffiliation{hpe}{Hewlett Packard Labs, Hewlett Packard Enterprise}
\mlsysaffiliation{msu}{Department of Computer Science and Engineering, Michigan State University}

\mlsyscorrespondingauthor{Yequan Zhao}{yequan\_zhao@ucsb.edu}
\mlsyscorrespondingauthor{Zheng Zhang}{zhengzhang@ece.ucsb.edu}

\mlsyskeywords{Machine Learning, MLSys}

\vskip 0.3in

\begin{abstract}
Physics intelligence and digital twins often require rapid and repeated performance evaluation of various engineering systems (e.g. robots, autonomous vehicles, semiconductor chips) to enable (almost) real-time actions or decision making. This has motivated the development of accelerated partial differential equation (PDE) solvers, in resource-constrained scenarios if the PDE solvers are to be deployed on the edge.  Physics-informed neural networks (PINNs) have shown promise in solving high-dimensional PDEs, but the training time on state-of-the-art digital hardware (e.g., GPUs) is still orders-of-magnitude longer than the latency required for enabling real-time decision making. Photonic computing offers a potential solution to address this huge latency gap  because of its ultra-high operation speed. However, the lack of photonic memory and the large device sizes prevent training real-size PINNs on photonic chips. This paper proposes a completely back-propagation-free (BP-free) and highly salable framework for training real-size PINNs on silicon photonic platforms. Our approach involves three key innovations: (1) a sparse-grid Stein derivative estimator to avoid the BP in the loss evaluation of a PINN, (2) a dimension-reduced zeroth-order optimization via tensor-train decomposition to achieve better scalability and convergence in BP-free training, and (3) a scalable on-chip photonic PINN training accelerator design using photonic tensor cores. We validate our numerical methods on both low- and high-dimensional PDE benchmarks. Through pre-silicon simulation based on real device parameters, we further demonstrate the significant performance benefit (e.g., real-time training, huge chip area reduction) of our photonic accelerator. Simulation codes will be released to the public after the paper review.
\end{abstract}
]



\printAffiliationsAndNotice{\mlsysEqualContribution} 

\section{Introduction} 

\subsection{Research Motivation}
Solving partial differential equations (PDE) is an indispensable step of high-fidelity modeling, simulation, verification and optimization of complex engineering design. 
Representative examples include, but are not limited to, electromagnetic and thermal analysis of IC chips~\citep{kamon1993fasthenry,li2004efficient}, computational fluid dynamic analysis for aircraft design~\cite{wang2014high}, medical imaging~\citep{villena2015marie}, and verification and control of autonomous systems~\citep{bansal2021deepreach}. Traditional numerical PDE solvers (\textit{e.g.,} finite-difference, finite-element methods) have been studied and optimized extensively on CPU and GPU, but their runtime often takes a few hours for practical engineering problems.
This runtime bottleneck becomes more significant in PDE-constrained inverse and control problems, since the forward problem needs to be solved many times in an outer iteration loop. 

As physical intelligence and digital twin surge, there have been increasing needs for solving PDEs with ultra-low latency (e.g., in real time) on edge computing platforms. This is often motivated by the real-time sensing, decision making and actions subject to physical laws and with limited computing and energy budget. For instance, ensuring safety is a top concern in self driving and autonomous systems. A classical safety verification approach computes the vehicle's reachable set by solving a high-dimensional Hamiltonian-Jacobi-Issac (HJI) PDE given the vehicle dynamics, uncertain environment and input disturbance~\citep{bansal2021deepreach}. Similarly, providing a safety-ensured control policy for proper operations of multi-agent robotic systems requires solving a high-dimensional Hamiltonian-Jacobi-Bellman (HJB) PDE~\cite{onken2021neural}.  In both applications, the PDE solutions are needed with ultra-low latency to ensure real-time decision making and control, which is far beyond the capability of existing PDE solvers. A similar challenge arise in the thermal management of modern digital processors (e.g., CPUs and GPUs)~\cite{liu2014parallel}. As massive heat is generated inside the high-density semiconductor chips, it is highly desirable to have a super fast built-in heat transfer PDE solver so that the processor itself can monitor and even control the internal temperature in real time. Unfortunately, current PDE-based thermal simulation tools supported via existing hardware platforms cannot meet this speed requirement. 

\subsection{Technical Challenges}
To fix the huge latency gap (i.e., hours of practical runtime versus real-time requirement) and resource constraints described above, we need significant advancement in both algorithm and hardware innovations. 

On the algorithm side, physics-informed neural networks (PINNs)~\citep{lagaris1998artificial,dissanayake1994neural,raissi2019physics} have emerged as a promising approach to solve both forward and inverse problems. Due to the discretization-free nature, PINN is more suitable for solving high-dimensional or parametric PDEs, such as the high-dimensional HJI/HJB PDEs that are beyond the capability of conventional discretization-based finite-difference or finite-element solvers. However, current PINN training is still too time consuming to deliver a low-latency solution. For example, training a PINN for robotic safety analysis and control~\citep{bansal2021deepreach,onken2021neural} can easily take $>10$ hours on a powerful GPU. This latency is far away from meeting the real-time requirement in practical autonomous systems. 
Despite the development of operator learning~\citep{lu2021learning}, a PINN often needs to be trained \textit{from scratch} again to obtain a high-quality solution once the PDE initial/boundary conditions, input functions or measurement data change.

On the hardware side, new hardware accelerators should be developed to train PINNs {\it ultra-fast} with constrained resources [e.g. limited hardware size, weight, and power (SWaP)]. However, the clock frequencies of conventional digital computing platforms (e.g., CPU and GPU) have hardly increased in the past 10 years due the physical limits of electronic chips and the end of Moore's Law~\cite{ross2008cpu}. Meanwhile, emerging computing platforms, such as integrated photonics, can offer a much higher clock frequency due to its ultra-high light operation speed~\cite{li2025all}. As a result, integrated photonic chips are considered a promising solution to achieve the ambitious goal of real-time sensing and scientific computing~\citep{NaPSAC} . 
In the past decade, many optical neural network (ONN) inference accelerators have been proposed~\citep{shen2017deep,tait2016microring,zhu2022space} for small-size neural networks.

However, designing a photonic training accelerator for real-size PINNs (e.g., a network with hundreds of neurons per layer) remains an open question due to two fundamental challenges.
\begin{itemize}
    \item {\bf Large device footprints and low integration density.} Photonic multiply-accumulate (MAC) units such as Mach-Zehnder interferometers (MZIs) are much larger ($\sim$10s of microns) than CMOS transistors. A real-size PINN with $>10^5$ parameters can easily exceed the available chip size with the square scaling rule, where a $N \times N$ weight matrix requires $O(N^2)$ MZIs ~\citep{triMZI,rectangleMZI}. In fact, even the state-of-the-art photonic AI {\it inference} accelerator~\citep{Ramey2020} can only handle $64\times 64$ weight matrices. Training a PINN on an photonic chip will face more significant scalability issue. 
    \item {\bf Difficulty of on-chip back propagation (BP).} It is hard to realize BP on photonic chips due to the lack of memory to store the computational graphs and intermediate results. Several BP-free and \textit{in-situ} BP methods~\citep{GuDAC, GuAAAI, filipovich2022silicon, buckley2022general, oguz2023forward,hughes2018training, pai2023experimentally} are proposed, but their scalability remains a major bottleneck. This becomes more severe in PINN, since its loss function also includes (high-order) derivative terms. Subspace learning~\citep{gu2021l2ight} may scale up BP-based training, but still needs to save intermediate states. 
\end{itemize}  
\vspace{-5pt}

BP-free training methods, especially stochastic zeroth-order optimization (ZO)~\citep{nesterov2017random, liu2020primer} or forward-forward method~\citep{hinton2022forward}, are easier to implement on edge hardware, since they do not need to detect or save any intermediate states. However, the scalability issue remains in {\it end-to-end} training, since the dimension-dependent gradient estimation error causes slow or even no convergence on PINNs with hundreds of neurons per layer. ZO training shows great success in fine-tuning large language models (LLMs)~\citep{malladi2023fine,yang2024adazeta,zhangrevisiting,gautamvariance}, since the gradient of a well-pretrained LLM has a low intrinsic dimension on fine-tuning tasks. Unfortunately, such a low-dimensional structure does not exist in end-to-end training, preventing the convergence of ZO optimization in training realistic PINNs. \citep{GuDAC,GuAAAI} utilized ZO training on a photonic chip, but it only fine-tuned a small portion of model parameters based on an offline pre-trained model.


\subsection{Research Contributions}
Different from the recent work of fine-tuning~\citep{GuDAC,GuAAAI,malladi2023fine,yang2024adazeta,zhangrevisiting,gautamvariance}, we investigate  {\bf end-to-end BP-free training of real-size PINNs on photonic chips from scratch}. This is a more challenging task because of (1) the differential operators in the PINN loss evaluation, and (2) the large number of optimization variables that cause divergence in end-to-end ZO training, (3) the scalability issue and lack of photonic memory on photonic chips.
This paper presents, for the first time, a {\bf real-size} and {\bf real-time} photonic accelerator to train PINNs with hundreds of neurons per layer on an integrated photonic platform. 

Our novel contributions are summarized as follows:
\begin{itemize}
\vspace{-5pt}
    \item{\bf Two-Level BP-free PINN Training.} We present a novel BP-free training approach in two implementation levels. Firstly, we propose a sparse-grid Stein estimator to calculate the (high-order) derivative terms in the PINN loss. Secondly, we propose a tensor-compressed variance reduction approach to improve the convergence of ZO-SGD. These innovations can completely bypass the need for photonic memory, and greatly improve the convergence of on-chip BP-free training. 
    \vspace{-5pt}
    \item{\bf A Scalable Photonic Design.} We present a scalable and easy-to-implement photonic accelerator design. 
    We reuse a tensorized ONN inference accelerator, and just add a digital controller to implement on-chip BP-free training. We present two designs: one implements the whole model on a single chip, and another uses a single photonic tensor core with time multiplexing. Our design can scale up to train real-size PINNs with hundreds of neurons per layer.
    \vspace{-5pt}
    \item {\bf Numerical Experiments and Hardware Emulation.} We validate our method in solving various PDEs. Our BP-free training achieves a competitive error compared to standard PINN training with BP, and achieves the lowest error compared with previous photonic on-chip training methods.
    We further evaluate the performance of our photonic training accelerator on solving a Black-Scholes PDE. The pre-silicon simulation results show that our design can reduce the number of MZIs by $42.7\times$, with only 1.64 seconds to solve this equation. 
\end{itemize} 
   
This work bridges the gap between physics-aware machine learning and hardware-aware system design, offering a general route toward real-time, low-power learning systems that can operate directly at the physical edge. 
Although this work does not provide chip fabrication results, the algorithmic results and pre-silicon evaluation provide sufficient evidence to show the great promise of our on-chip learning framework in the coming era of physical intelligence and digital twins.   


\section{Background}

\paragraph{Physics-Informed Neural Networks (PINNs).}
Consider a generic PDE:
\begin{equation}
\begin{aligned}
\mathcal{N}[\bm{u}(\bm{x},t)]&=l(\bm{x}, t), ~\quad \bm{x} \in \Omega,~~t \in[0, T],\\
\mathcal{I}[\bm{u}(\bm{x},0)] &= g(\bm{x}), ~~~\quad \bm{x} \in \Omega,\\
\mathcal{B}[\bm{u}(\bm{x},t)] &= h(\bm{x}, t),  \quad \boldsymbol{x} \in \partial \Omega, ~~t \in[0, T],  
\end{aligned}
\label{general PDE}
\end{equation}
where $\bm{x}$ and $t$ are the spatial and temporal coordinates; $\Omega \subset \mathbb{R}^{D}$, $\partial \Omega$ and $T$ denote the spatial domain, domain boundary and time horizon, respectively; $\mathcal{N}$ is a nonlinear differential operator; $\mathcal{I}$ and $\mathcal{B}$ represent the initial and boundary condition; $\bm{u} \in \mathbb{R}^{n}$ is the solution for the PDE described above. In the contexts of PINNs~\citep{raissi2019physics}, a solution network $\bm{u}_{\bm{\theta}}(\bm{x},t)$, parameterized by $\bm{\theta}$, is substituted into PDE \eqref{general PDE}, resulting in a residual defined as:
\begin{equation}
r_{\bm{\theta}}(\bm{x},t):=\mathcal{N}[\bm{u}_{\bm{\theta}}(\bm{x},t)]-l(\bm{x}, t).
\label{PDE residual}
\end{equation}
The parameters $\bm{\theta}$ can be trained by minimizing the loss:
\begin{equation}
\mathcal{L}(\bm{\theta})=\mathcal{L}_r(\bm{\theta})+\lambda_{0}\mathcal{L}_0(\bm{\theta})+\lambda_{b}\mathcal{L}_b (\bm{\theta}).
\label{PINNs loss}
\end{equation}
Here $\mathcal{L}_r (\bm{\theta})$, $\mathcal{L}_0 (\bm{\theta})$ and $\mathcal{L}_b (\bm{\theta})$ are the residuals associated with the PDE operator, the initial condition and boundary condition, respectively. The residual of PDE operator
\begin{equation}
\begin{aligned}
\mathcal{L}_r (\bm{\theta}) &= \frac{1}{N_{r}}\sum_{i=1}^{N_{r}}  \left\|r_{\bm{\theta}}(\bm{x_{r}}^{i},t_{r}^{i})\right\|_{2}^{2}, 
\end{aligned}
\label{loss terms}
\end{equation}
involves (high-order) derivative terms.

\paragraph{Zeroth-Order (ZO) Optimization.}
We consider minimizing a loss function $\mathcal{L}(\bm{\theta})$ by updating  parameters $\bm{\theta} \in \mathbb{R}^d$ iteratively using a (stochastic) gradient descent method: 
\begin{equation}
    \bm{\theta}_t \leftarrow \bm{\theta}_{t-1}-\alpha \bm{g}
    \label{SGD update}
\end{equation}
where $\bm{g}$ denotes the (stochastic) gradient of the loss $\mathcal{L}$ w.r.t. model parameters $\bm{\theta}$. ZO optimization uses a few forward function queries to approximate the gradient $\bm{g}$:
\begin{equation}
   \bm{g} \approx \hat{\nabla}_{\bm{\theta}}\mathcal{L}(\bm{\theta})=
    \sum_{i=1}^N \frac{1}{2N\mu} \left[\mathcal{L}\left(\bm{\theta}+\mu \bm{\xi}_i\right)-\mathcal{L}(\bm{\theta}-\mu \bm{\xi}_i)\right] \bm{\xi}_i.
\label{ZO gradient estimation}
\end{equation}
Here $\{\bm{\xi}_i\}_{i=1}^N$ are some perturbation vectors and $\mu$ is the sampling radius, which is typically small. We consider the random gradient estimator (RGE), in which $\{\bm{\xi}_i\}_{i=1}^N$ are $N$ i.i.d. samples drawn from a distribution $\rho(\bm{\xi})$ with zero mean and unit variance. 
The variance of RGE involves a dimension-dependent factor $O(d/N)$ given $\mu = O(1/\sqrt{N})$~\citep{liu2020primer}. ZO optimization has been used extensively in signal processing and adversarial machine learning \citep{ZOSGD, duchi2015optimal, ZOSCD, chen2019zo,shamir2017optimal,cai2021zeroth}. Recently, ZO optimization has achieved great success in fine-tuning LLMs~\citep{malladi2023fine,yang2024adazeta,zhangrevisiting,gautamvariance}, due to the low intrinsic dimensionality of the gradient information. Without low-dimensional structures, ZO optimization scales poorly in end-to-end training of real-size neural networks due to the large dimension-dependent gradient variance. Recently, \citep{chen2023deepzero} improved the scalability of end-to-end ZO training by exploiting model sparsity, but its coordinate-wise gradient estimation is prohibitively expensive for edge devices or real-time applications.

\paragraph{Optical Neural Networks (ONN) and On-chip ONN Training.} Photonic AI accelerators are expected to outperform their electronic counterparts due to low latency, ultra-high throughput, high energy efficiency, and high parallelism~\citep{mcmahon2023physics}. 
Due to limited scalability, state-of-the-art photonic AI accelerators can only handle weight matrices of size 64$\times$64~\citep{Ramey2020}. As a result, large-scale optical matrices are computed by tiles or blocks with time multiplexing, requiring intensive memory access to store the intermediate data.
\cite{demirkiran2023electro} shows that only $\sim$10\text{\%} of the overall power is consumed in optical devices. Applying a pre-trained model on non-ideal photonic chips usually faces significant performance degradation. On-chip training is essential to mitigate this degradation. 
However, there is no access to intermediate states or full gradients on the photonic chip. The existing BP-based method \cite{hughes2018training, wright2022deep, pai2023experimentally} requires external hardware to perform gradient computation, which is bulky and not scalable. Several BP-free methods are proposed to address this issue \citep{GuDAC, GuAAAI, filipovich2022silicon, buckley2022general, oguz2023forward}. However, these methods can only handle a small number of training parameters.

\section{Two-Level BP-free Training for PINNs}
Current PINN training methodologies rely on BP for both loss evaluations [Eq. (\ref{PINNs loss})] and model parameter updates [Eq. (\ref{SGD update})]. These BP computations are hard to implement on photonic chips. This section proposes a two-level BP-free PINN training framework to avoid such a challenge. 
This approach improves the convergence of the training framework and the scalability on photonic chips, enabling {\it end-to-end} training of real-size PINNs with hundreds of neurons per layer.


\subsection{Level 1: BP-Free PINN Loss Evaluation}

\subsubsection{Stein Derivative Estimation}
For an input $\bm{x}\in \mathbb{R}^{D}$ and an approximated PDE solution $\bm{u}_{\bm{\theta}}(\bm{x})\in \mathbb{R}^{n}$ parameterized by $\bm{\theta}$, we consider the first-order derivative $\nabla_{\bm{x}} \bm{u}_{\bm{\theta}}$ and Laplacian $\Delta \bm{u}_{\bm{\theta}}$ involved in the loss function of a PINN training. Our implementation leverages the Stein  estimator~\citep{stein1981estimation}. Specifically, we represent the PDE solution $\bm{u}_{\bm{\theta}}(\bm{x})$ via a Gaussian smoothed model:
\begin{equation}
\bm{u}_{\bm{\theta}}(\bm{x})=\mathbb{E}_{\bm{\delta} \sim \mathcal{N}\left(\bm{0}, \sigma^2 \bm{I}\right)} f_{\bm{\theta}}(\bm{x}+\bm{\delta}),
\label{gaussian smoothed model}
\end{equation}
where $f_{\bm{\theta}}$ is a neural network with parameters $\bm{\theta}$; $\bm{\delta} \in \mathbb{R}^{D}$ is the random noise sampled from a multivariate Gaussian distribution ${\cal N}(\bm{0}, \sigma^2 \bm{I})$. The first-order derivative and Laplacian of $\bm{u}_{\bm{\theta}}(\bm{x})$ can be written as:
\begin{equation}
\begin{aligned}
 \nabla_{\bm{x}} \bm{u}_{\bm{\theta}}=\mathbb{E}_{\bm{\delta} \sim \mathcal{N}\left(\bm{0}, \sigma^2 \bm{I}\right)} & \left[\frac{\bm{\delta}}{2 \sigma^2}(f_{\bm{\theta}}(\bm{x}+\bm{\delta}) - f_{\bm{\theta}}(\bm{x}-\bm{\delta}))\right],\\
\Delta \bm{u}_{\bm{\theta}}=\mathbb{E}_{\bm{\delta} \sim \mathcal{N}\left[\bm{0}, \sigma^2 \bm{I}\right)}  & \left [f_{\bm{\theta}}(\bm{x}+\bm{\delta}) +
f_{\bm{\theta}}(x-\bm{\delta})-2 f_{\bm{\theta}}(\bm{x})\right]\\
& \times \frac{\|\bm{\delta}\|^2-\sigma^2 D}{2 \sigma^4}.  
\end{aligned}
\label{stein derivative estimator}
\end{equation}
In~\cite{he2023learning}, the above expectation is computed by evaluating $f_{\bm{\theta}}(\bm{x}+\bm{\delta})$ and $f_{\bm{\theta}}(\bm{x}-\bm{\delta})$ at a set of i.i.d. Monte Carlo samples of $\bm{\delta}$. However, Monte Carlo needs massive (e.g., $>10^3$) function queries. Therefore, it is highly desirable to develop a more efficient BP-free method for evaluating derivative terms in the loss function. 

\subsubsection{Sparse-Grid Stein Derivative Estimator}
Now we use the sparse grid techniques~\citep{garcke2006sparse,gerstner1998numerical} to significantly reduce the number of function queries in the Stein derivative estimator, while maintaining high accuracy in numerical integration. This approach has been widely used in uncertainty quantification, but has not been utilized for training PINNs.

We begin with a sequence of univariate quadrature rules $V=\left\{V_l: l \in \mathbb{N}\right\}$. Here $l$ denotes an accuracy level so that any polynomial function of order $\leq l$ can be exactly integrated with $V_l$. Each rule $V_l$ specifies $n_{l}$ nodes $N_{l}=\left\{\delta_{1},\dots,\delta_{n_{l}}\right\}$ and weight function $w_l: N_{l} \rightarrow \mathbb{R}$. With $V_{k}$, the integration a $f$ over a random variable $\delta$ is written as:
\begin{equation}
\int_{\mathbb{R}} f(\delta) p(\delta) \, d\delta \approx 
V_{k}[f]=\sum_{\delta_j \in N_{k}} w_{k}(\delta_j) f(\delta_j).
\end{equation}
Here $p(\delta)$ is the probability density function (PDF) of $\delta$. 

Next, we consider the multivariate integration of a function $f$ over a random vector $\bm{\delta}=(\delta^{1},\dots,\delta^{D})$, where $p(\bm{\delta})=\prod_{m=1}^{D}p(\delta^{m})$ is the joint PDF. Let the multi-index $\bm{l} = (l_1, l_2, ..., l_D) \in \mathbb{N}^{D}$ specify the desired integration accuracy for each dimension. We use the Smolyak algorithm~\citep{gerstner1998numerical} to construct sparse grids. For any non-negative integer $q$, define $\mathbb{N}_{q}^{D} = \left\{\bm{l}\in \mathbb{N}^{D}: \sum_{m=1}^{D}l_{m} = D+q\right\}$ and $\mathbb{N}_{q}^{D} = \emptyset$ for $q<0$. The level-$k$ Smolyak rule $A_{D,k}$ for $D$-dim integration can be written as~\citep{wasilkowski1995explicit}:
\small
\begin{equation}
\begin{aligned}
A_{D,k}[f] =    \sum_{q=k-D}^{k-1} & (-1)^{k-1-q}\left(\begin{array}{c}
D-1 \\ k-1-q
\end{array}\right)  \times \\
& \sum_{\bm{l} \in \mathbb{N}_q^D}       \left(V_{l_{1}} \otimes \cdots \otimes V_{l_{D}} \right)[f] .
\end{aligned}
\end{equation}\normalsize
It follows that:
\begin{equation}
\begin{aligned}
A_{D,k}[f] = \sum_{q=k-D}^{k-1} \sum_{\bm{l} \in \mathbb{N}_q^D} \sum_{\delta^{1} \in N_{l_{1}}} \cdots \sum_{\delta^{D} \in N_{l_{D}}} (-1)^{k-1-q} \times\\
\left(\begin{array}{c}
D-1 \\ k-1-q
\end{array}\right)  \prod_{m=1}^{D} w_{l_{m}}(\delta^{m})f(\delta^{1},\dots,\delta^{D}), \nonumber
\end{aligned}
\end{equation}
which is a weighted sum of function evaluations $f(\bm{\delta})$ for $\bm{\delta} \in \bigcup_{q=k-D}^{k-1} \bigcup_{\mathbf{l} \in \mathbb{N}_q^D} \left(N_{l_{1}} \times \cdots \times N_{l_{D}}\right)$. 
For the same $\bm{\delta}$ that appears multiple times for different combinations of values of $\bm{l}$, we only need to evaluate $f$ once and sum up the respective weights beforehand. The resulting level-$k$ sparse quadrature rule defines a set of $n_{L}$ nodes $S_{L}=\left\{\bm{\delta}_{1},\dots,\bm{\delta}_{n_{L}}\right\}$ and the corresponding weights $\left\{w_{1},\dots,w_{n_{L}}\right\}$. The $D$-dim integration can then be efficiently computed with the sparse grids as:
\begin{equation}
\int_{\mathbb{R}^{D}}f(\bm{\delta})p(\bm{\delta}) d \bm{\delta} \approx A_{D,k}[f] = \sum_{j=1}^{n_{L}} w_{j}f(\bm{\delta}_{j}).
\end{equation}
In practice, since the sparse grids and the weights do not depend on $f$, they can be pre-computed for the specific quadrature rule, dimension $D$, and accuracy level $k$. 

Finally, we implement the Stein derivative estimator in Eq. (\ref{stein derivative estimator}) via the sparse-grid integration. Noting that $\bm{\delta} \sim \mathcal{N}\left(\bm{0}, \sigma^2 \bm{I}\right)$, we can use univariate Gaussian quadrature rules as basis to construct a level-$k$ sparse Gaussian quadrature rule $A_{D,k}^{*}$ for $D$-variate integration. Then the first-order derivative and Laplacian in Eq. (\ref{stein derivative estimator})  is approximated as:
\small
\begin{equation}
\begin{aligned}
\nabla_{\bm{x}} \bm{u}_{\bm{\theta}}& \approx  \sum_{j=1}^{n_{L}^{*}}w_{j}^{*}  \left[\frac{\bm{\delta}_{j}^{*}}{2 \sigma^2}(f_{\bm{\theta}}(\bm{x}+\bm{\delta}_{j}^{*}) - f_{\bm{\theta}}(\bm{x}-\bm{\delta}_{j}^{*}))\right],  \\
\Delta \bm{u}_{\bm{\theta}}  & \approx \sum_{j=1}^{n_{L}^{*}}w_{j}^{*} \left(\frac{\|\bm{\delta}_{j}^{*}\|^2-\sigma^2 D}{2 \sigma^4}\right) \times \\
& \left( f_{\bm{\theta}}(\bm{x}+\bm{\delta}_{j}^{*}) +
f_{\bm{\theta}}(x-\bm{\delta}_{j}^{*})-2 f_{\bm{\theta}}(\bm{x}) \right),
\end{aligned}
\label{sparse-grid stein derivative estimator}
\end{equation} \normalsize
where $\bm{\delta}_{j}^{*}$ and $w_{j}^{*}$ are defined by the sparse grid $A_{D,k}^{*}$. 

\begin{figure*}[!t]
    \centering
    \includegraphics[width=0.9\linewidth]{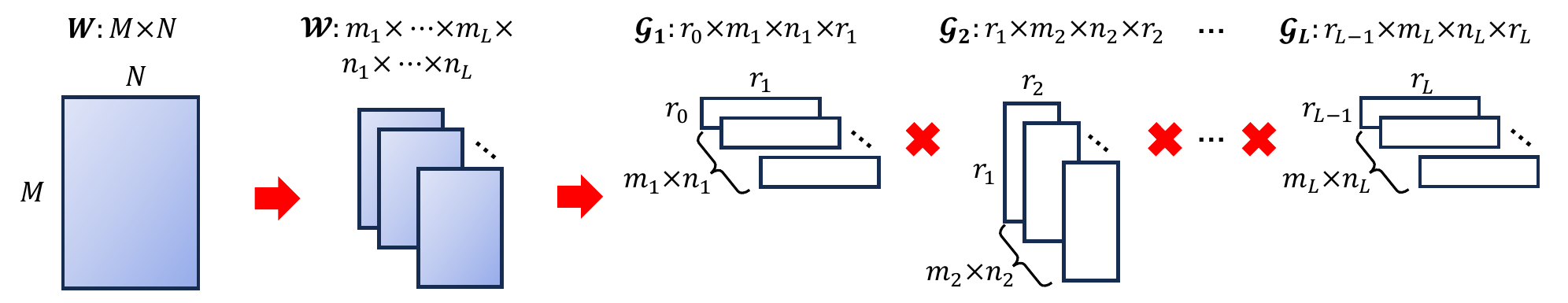}
    \caption{Tensor-train decomposition: matrix $\mat{W}$ is folded to a multi-way tensor $\ten{W}$ and decomposed into $L$ small TT cores $\{\ten{G}_k\}_{k=1}^L$.}
    \label{fig:tensor_train}
\end{figure*}

\paragraph{Remark:} With the sparse-grid Stein estimator in Eq. (\ref{sparse-grid stein derivative estimator}), we can compute the derivatives in Eq. (\ref{PDE residual}) and the loss of PINN in Eq. (\ref{PINNs loss}) without using any BP computation. Recent work explored efficient computation of differential operators \cite{cho2024separable,shi2024stochastic}, however, they still need the automatic differentiation, which is not available for optical neural networks. Our BP-free sparse-grid loss evaluation offers two benefits:
\begin{itemize}
\vspace{-5pt}
    \item The number of forward evaluations (i.e. $n_{L}^{*}$) is usually significantly smaller than the number of Monte Carlo samples required to evaluate Eq. (\ref{stein derivative estimator}). For example, a level-3 sparse-grid Gaussian quadrature for a 3-dim PDE requires only 25 function evaluations, compared to thousands in Monte Carlo-based estimation~\cite{he2023learning}.
    \vspace{-5pt}
    \item The sparse-grid Stein estimator transforms differential operators into the weighted sums of forward evaluations. This often leads to a smoother loss function and thus a better generalization accuracy~\cite{wen2018smoothout}. This will be empirically shown in Section~\ref{subsec:numerical}.
\end{itemize}

\subsection{Level 2: Tensor-Compressed ZO Training}\label{subsec:TT}

To avoid BP in the update of PINN model parameters, we use the ZO gradient estimator in Eq (\ref{ZO gradient estimation}) to perform gradient-descent iteration. Considering the inquiry complexity, we use randomized gradient estimation to implement (\ref{ZO gradient estimation}). The gradient mean squared approximation error scales with the perturbation dimension $d$~\citep{berahas2022theoretical}:
$
    \mathbb{E}\left[\|\hat{\nabla}_{\bm{\theta}}\mathcal{L}(\bm{\theta}) - \nabla_{\bm{\theta}}\mathcal{L}(\bm{\theta}) \|_2^2 \right] = O\left(\frac{d}{N}\right)\|\nabla_{\bm{\theta}}\mathcal{L}(\bm{\theta})\|_2^2+O\left(\frac{\mu^2 d^3}{N}\right)+O\left(\mu^2 d\right)
$.
The convergence rate also scales with $d$ as $O(\sqrt{d}/\sqrt{T})$ in non-convex unconstrained optimization~\citep{berahas2022theoretical}.
Real-size PINNs typically have hundreds of neurons per hidden layer, and the total number of model parameters can easily exceed $10^5$ or $10^6$.
As a result, ZO optimization converges slowly or even fails to converge in {\it end-to-end} PINN training. 
      
\subsubsection{Tensor-Compressed ZO Optimization.} We propose to significantly reduce the gradient variance via a \textit{low-rank} tensor-compressed training as shown in Fig.~\ref{fig:tensor_train}. Tensor compression has been well studied for functional approximation and data/model compression~\cite{lubich2013dynamical,zhang2016big,novikov2015tensorizing}, but it has not been studied for variance reduction in ZO training.

Let $\bm{W} \in \mathbb{R}^{M\times N}$ be a weight matrix in a PINN. We factorize its dimension sizes as $M = \prod^{L}_{i=1}m_i$ and $N = \prod^{L}_{j=1}n_j$, fold $\bm{W}$ into a $2L$-way tensor $\mathbfcal{W} \in \mathbb{R}^{m_1\times m_2 \times \dots \times m_L \times n_1 \times n_2 \times \dots \times n_L}$, and write $\ten{W}$ with the tensor-train (TT) decomposition \citep{oseledets2011tensor}:
\begin{equation}
{\mathbfcal{W}}(i_1, i_2, \dots, i_L, j_1, j_2, \dots, j_L)
\approx \prod^{L} \limits_{k=1} \mat{G}_k(i_k, j_k)
\end{equation}
Here $\mat{G}_k(i_k, j_k) \in \mathbb{R}^{r_{k-1} \times r_{k}}$ is the $(i_k, j_k)$-th slice of the TT-core $\mathbfcal{G}_k \in \mathbb{R}^{r_{k-1}\times m_k \times n_k \times r_k}$ by fixing its $2$nd and $3$rd indices as $i_k$ and $j_k$, respectively. The vector $(r_0, r_1, \dots, r_{L})$ is called the TT-ranks with $r_0=r_{L}=1$. TT representation reduces the number of variables from $\prod_{k=1}^{L} m_k n_k$ to $\sum_{k=1}^{L}r_{k-1}m_k n_k r_k$. The compression ratio is controlled by the TT-ranks, which can be learnt automatically~\citep{hawkins2021bayesian,hawkins2022towards}. 

In ZO training, we directly train the TT factors $\{\mathbfcal{G}_k \}_{k=1}^L$. 
Take a weight matrix with size $512\times 512$ for example, the original dimension $d=2.62\times 10^5$, while the reduced number of variables in TT factors is $d'=256$ (with tensor size $8\times 4\times 4\times 4 \times 4 \times 4\times 4\times 8$, and TT-rank (1,2,2,2,1)).
This reduces the problem dimensionality $d$ by $1023\times$, leading to dramatic {\it variance reduction} of the ZO gradient in Eq. (\ref{ZO gradient estimation}). As will be shown in Table \ref{tab:weight training}, such dimension reduction does little harm to the model learning capacity, but greatly improves the ZO training convergence.


\paragraph{Comparison with other ZO Training.} Other techniques have also been reported to improve the convergence of ZO training, such as sparse ZO optimization~\citep{chen2023deepzero,liu2024sparse} and ZO variance-reduced gradient descent (SVRG)~\citep{liu2018zeroth}. Although these techniques can improve the convergence of ZO training, they cannot reduce the hardware complexity (i.e., the number of photonic devices needed for hardware implementation). The ZO SVRG method needs storing previous gradient information, thus can cause huge memory overhead and is not suitable for photonic implementation. Our method, as will be shown in Section~\ref{label: photonic design}, can improve both the convergence and scalability of photonic training. 




\section{Design with Integrated Photonics}\label{label: photonic design}

\begin{figure*}[t]
    \centering
    \includegraphics[width=0.9\textwidth]{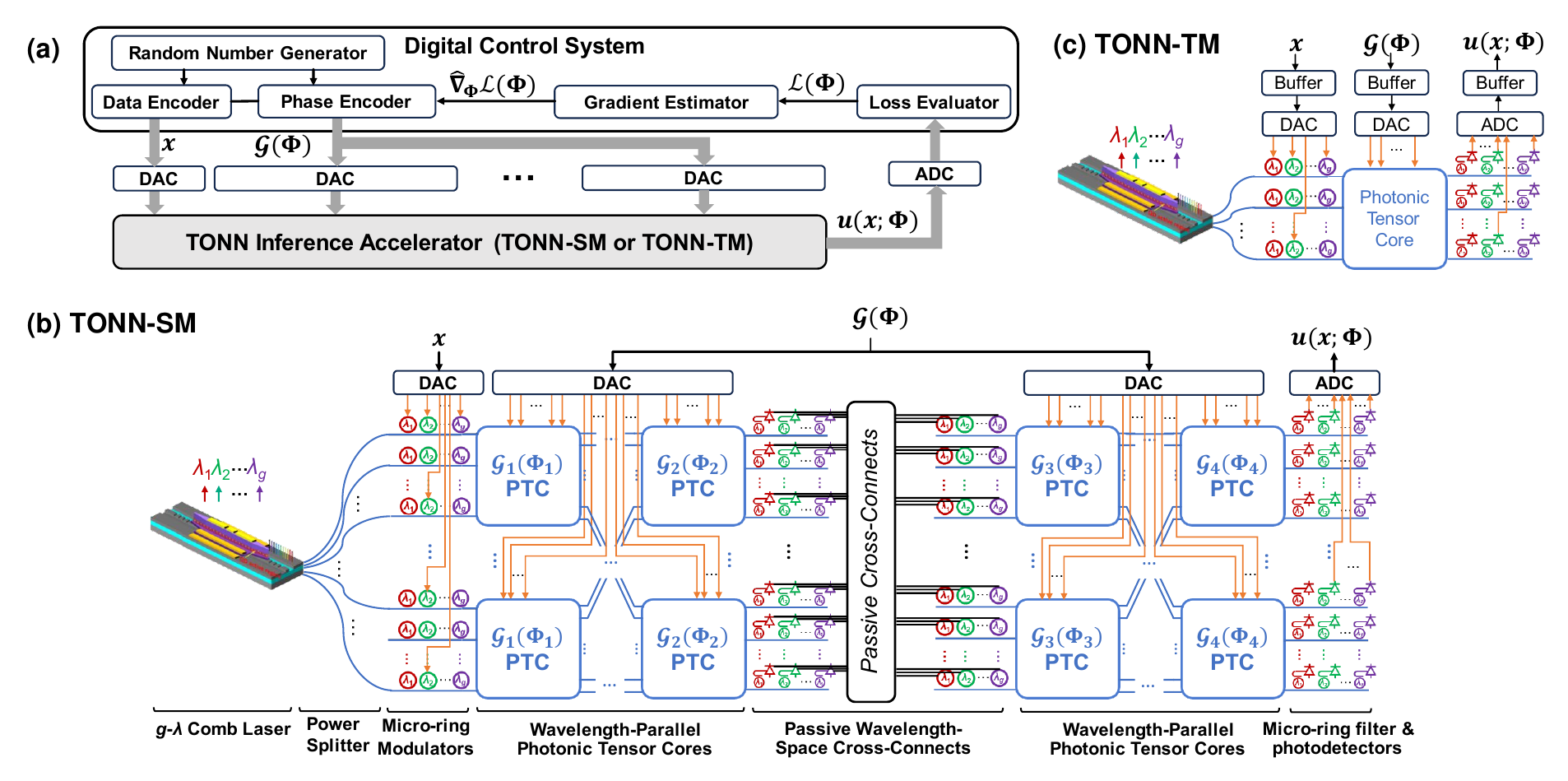}
    \vspace{-10pt}
    \caption{(a) The overall architecture of the BP-free optical training accelerator. (b) TONN space multiplexing (TONN-SM) architecture. (c) TONN time multiplexing (TONN-TM) architecture.}
    \label{fig:TONN}
\end{figure*}

This section presents the design of our photonic PINN training accelerator. Due to the BP-free nature, we can reuse a photonic inference accelerator to easily finish the training hardware design. The tensor-compressed ZO training can greatly reduce the number of required photonic devices, providing much better scalability than existing work. 

    \vspace{-5pt}
    \paragraph{Overall Architecture.}
    Figure \ref{fig:TONN} (a) shows the architecture of our optical PINN training accelerator. The accelerator consists of an ONN inference engine and a digital control system to implement BP-free PINN training. As explained in Appendix~\ref{appendix:ONN Basics}, standard ONN~\citep{shen2017deep} architecture uses singular value decomposition (SVD) to implement matrix-vector multiplication (MVM), and unitary matrices are implemented with MZI meshes~\citep{rectangleMZI}. For a $N\times N$ weight matrix, this requires $O(N^2)$ MZIs, which is {\it infeasible} for practical PINNs. In contrast, our method utilizes tensor-compressed ZO training, therefore we utilize the tensorized ONN (TONN) accelerator~\citep{xiao2021large} as our inference engine. A TONN inference accelerator only implements the photonic TT cores $\{\ten{G}_k\}_{k=1}^L$ instead of the matrix $\mat{W}$ on an integrated photonic chip, greatly reducing the number of MZIs required for large-scale implementation. 
    The target of on-chip ONN training is to find the optimal MZI phases $\bm{\Phi}$ under various variations. 
    We implement BP-free PINN training by updating the MZI phases $\bm{\Phi}_k$ in each photonic TT core $\ten{G}_k(\bm{\Phi}_k)$. 
    

    \vspace{-5pt}
    
    \paragraph{Two Tensorized ONN (TONN) Inference Accelerators.}
    Here we present two designs for the TONN inference. 
    The space multiplexing design (TONN-SM in Fig.~\ref{fig:TONN} (b)) integrates the whole tensor-compressed model on a single chip. Each TT core is implemented by several identical photonic tensor cores.
    Tensor multiplications between input data and all TT-cores are realized in a single clock cycle by cascading the photonic TT-cores in the space domain and adding parallelism in the wavelength domain \citep{xiao2021large}. 
    The time multiplexing design (TONN-TM in Fig.~\ref{fig:TONN}(c)) uses a single wavelength-parallel photonic tensor core~\citep{Xiao2023}. In each clock cycle, the photonic tensor core with parallel processing in the wavelength domain is updated to multiply with the input tensor. The intermediate output data are then stored in the buffer for the next cycle. 
    TONN-SM is fast and ``memory-free". TONN-TM exhibits a smaller footprint at the cost of higher latency and additional memory requirements.
    We provide more details in Appendix \ref{appendix:TONN Implementation Details}.


    \paragraph{BP-free On-chip PINN Training.} 
    BP-free training repeatedly calls a TONN inference engine to evaluate the loss and estimate the gradients, then update the MZI phases. 
    To get the ZO gradient $\hat{\nabla}_{\boldsymbol{\Phi}}\mathcal{L}(\boldsymbol{\Phi})$ given by Eq. (\ref{ZO gradient estimation}), the digital control system generates Rademacher random perturbations (entries are integers +1 or -1 with equal probability) and re-program the MZIs with the perturbed phase values $\boldsymbol{\Phi}+\mu \boldsymbol{\xi}$. Here we set $\mu$ as the minimum control resolution of MZI phase tuning. Loss evaluation $\mathcal{L}(\bm{\Phi}+\mu \boldsymbol{\xi})$ requires a few inferences with perturbed input data to estimate first- and second-order derivatives by sparse-grid Stein estimator.
    The digital controller gathers the gradient estimation of $N$ i.i.d. perturbations, and update the MZI phases with $\hat{\nabla}_{\boldsymbol{\Phi}}\mathcal{L}(\boldsymbol{\Phi})$.

    
\section{Experimental Results}
To validate our method, we consider 4 PDE benchmarks: (1) a 1-dim Black-Scholes equation modeling call option price dynamics in financial markets, (2) a 20-dim Hamilton-Jacobi-Bellman (HJB) equation arising from optimal control of robotics and autonomous systems, (3) a 1-dim Burgers' equation \cite{hao2023pinnacle}, (4) a 2-dim Darcy Flow problem \cite{li2020fourier}. Detailed PDE formulations are given in Appendix \ref{appendix:PDE details}. The baseline neural networks are: 3-layer MLPs with 128 neurons per layer and \texttt{tanh} activation for Black-Scholes, 3-layer MLP with 512 neurons per layer and \texttt{sine} activation for 20-dim HJB, 5-layer MLPs with 100 neurons per layer and \texttt{tanh} activation for Burgers' equation and the Darcy flow problem. 

The PINN models are trained using Adam optimizer (learning rate 1e-3), implementing first-order (FO) and ZO training approaches. FO training uses true gradients computed by BP, whereas ZO training uses RGE gradient estimation. For ZO training, we set the query number $N=1$, smoothing factor $\mu=0.01$, and use tensor-wise gradient estimation.
We evaluate model accuracy on a hold-out set using the relative $\ell_2$ error $\|\hat{u}-u\|^2 / \|u \|^2$ in domain $\Omega$, where $\hat{u}$ is the model prediction and $u$ is the reference solution. We repeat all experiments three times and record the mean values and standard deviations. See details in Appendix \ref{appendix:Training Set-ups}.

\subsection{Numerical Results of Solving Various PDEs}
\label{subsec:numerical}
We first evaluate the numerical performance of our BP-free PINNs training algorithm. We conduct training in the \textit{weight domain}, where the trainable parameters are the weight matrices $\mat{W}$ (tensor cores $\ten{G}$ in tensor-compressed training) with tractable gradients to enable FO training as baselines.

\vspace{-10pt}

\begin{table}[t]
    \begin{minipage}{\linewidth}
        \centering
        \caption{Relative $\ell_2$ error of different loss computation methods.}
        \begin{tabular}{c|ccc}
        \toprule
        \toprule
        Problem & AD    & SE    & SG (ours) \\
        \midrule
        Black-Scholes & 5.35E-02 & 5.41E-02 & \textbf{5.28E-02} \\
        20-dim HJB & 1.99E-03 & 1.52E-03 & \textbf{8.16E-04} \\
        Burgers & 1.37E-02 & 1.98E-02 & \textbf{1.31E-02} \\
        Darcy Flow & 7.57E-02 & 7.85E-02 & \textbf{7.47E-02} \\
        \bottomrule
        \bottomrule
        \end{tabular}%
             
        \label{tab:loss computation}%
    \end{minipage}

    \begin{minipage}{\linewidth}
        \centering
        \caption{Relative $\ell_2$ error of different training methods. }
        \resizebox{\linewidth}{!}{
        \begin{tabular}{c|cc|cc}
        \toprule
        \toprule
        Problem & \multicolumn{2}{c|}{FO Training} & \multicolumn{2}{c}{ZO Training} \\
        \cmidrule{2-5}      & Standard & TT    & Standard & TT \\
        \midrule
        Black-Scholes & \textbf{5.28E-02} & 5.97E-02 & 3.91E-01 & \textbf{8.30E-02} \\
        20-dim HJB & 8.16E-04 & \textbf{2.05E-04} & 6.86E-03 & \textbf{1.54E-03} \\
        Burgers & \textbf{1.31E-02} & 4.49E-02 & 4.41E-01 & \textbf{1.63E-01} \\
        Darcy Flow & \textbf{7.47E-02} & 8.77E-02 & 1.34E-01 & \textbf{9.05E-02} \\
        \bottomrule
        \bottomrule
        \end{tabular}%

        }
        \label{tab:weight training}%
    \end{minipage}
   \vspace{-10pt}
\end{table}

\paragraph{Effectiveness of BP-free Loss Computation:} We consider three methods for computing derivatives in the loss of PINN: 1) BP via automatic differentiation (\textbf{AD}) as a gold reference, 2) BP-free Monte Carlo Stein Estimator (\textbf{SE})~\citep{he2023learning} using 2048 random samples, and 3) our proposed BP-free method via sparse-grid (\textbf{SG}). Details are provided in Appendix \ref{appendix:Loss Evaluation Set-ups}.
We perform FO training on standard PINNs for fair comparison. As shown in Table \ref{tab:loss computation}, 
our proposed \textbf{SG} outperforms {\bf SE} while requiring much less forward evaluations. In most cases \textbf{SG} outperforms \textbf{AD}. We hypothesize that this is attributed to the smoothed loss that {\it improves generalization}~\cite{wen2018smoothout}. 




\paragraph{Evaluation of BP-free PINN Training.} We compare the \textbf{FO} training (BP) and \textbf{ZO} training (BP-free) in the standard (\textbf{Std.}) uncompressed and our tensor-train (\textbf{TT}) compressed formats. We employ {\bf SG} loss computation for all experiments. Table \ref{tab:weight training} summarizes the results. 
\textbf{TT dimension-reduction does little harm to the accuracy of the PINN model:}
Tensor-compressed training reduces the dimension by $20.44\times, 142.27\times, 24.74\times, 24.74\times$ for Black-Scholes, 20-dim HJB, Burgers', and Darcy Flow, respectively.
The first two columns list the relative $\ell_2$ error achieved after FO training. 
TT compressed training achieves an error similar to standard training with FC hidden layers. 
\textbf{TT dimension reduction greatly improves the convergence of ZO training:}
The last two columns list the relative $\ell_2$ error achieved after ZO training. Standard ZO training fails to converge well due to the high gradient variance which stems from the high dimensionality.
Using TT to reduce the variance of the gradient, our ZO training method achieves much better convergence and final accuracy. 
This showcases that our proposed TT compressed ZO optimization is the key to the success of BP-free training on real-size PINNs. 
The observations above clearly demonstrate that our method can bypass BP in both loss evaluation and model parameter updates, and is still capable of learning a good solution.
We defer the error curves and average results of repeated experiments to Appendix \ref{appendix:Detailed Results of Weight-domain Training}. 

\vspace{-10pt}
\paragraph{Ablation Studies.} We performed ablation studies to assess the impact of architectural and algorithmic design choices, including TT-rank selection, hidden-layer width, sparse-grid derivative estimation, and hyperparameters in tensor-compressed ZO optimization. The results show that reducing hidden-layer width increases test errors, confirming that sufficient model capacity is necessary and our baseline networks are not over-parameterized. A small TT-rank ($r=2$) achieves near-optimal accuracy with high hardware efficiency, while larger ranks yield marginal gains but higher complexity. The sparse-grid Stein estimator converges faster and attains lower test error than the Monte Carlo variant using far fewer samples. The chosen hyperparameters for sparse-grid estimation and ZO optimization balance convergence speed and final accuracy. Detailed results are provided in Appendix \ref{appendix:Ablation Studies}.

\begin{figure}[t]
    \centering
    \includegraphics[width=\linewidth]{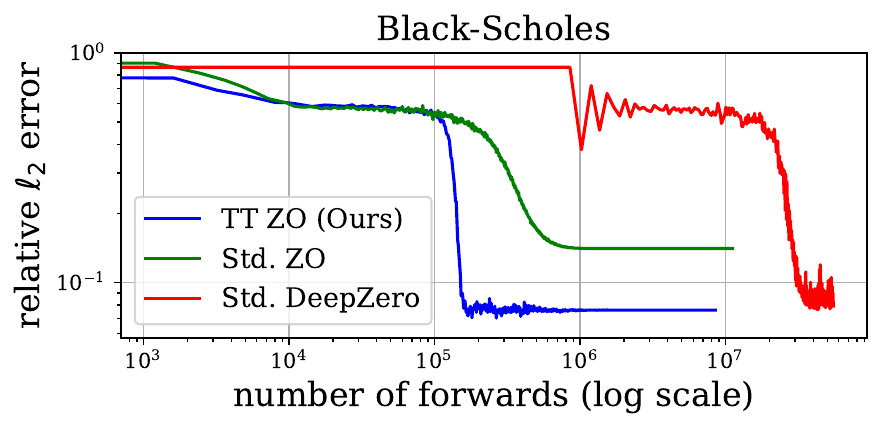}
    \vspace{-20pt}
    \caption{Training efficiency comparison of ZO training methods.}
    \label{fig:deep zero}
    \vspace{-10pt}
\end{figure}

\vspace{-10pt}
\paragraph{Remark.} 
There is an avoidable performance gap between FO training and all ZO training, due to the additional variance term of ZO gradient estimation. While this gap cannot be completely eliminated, it may be narrowed by using more forward passes per iteration in the late training stage to achieve a low-variance ZO gradient [\textit{e.g.,} ZO-RGE with a large N or coordinate-wise gradient estimator used in \texttt{DeepZero}~\citep{chen2023deepzero}]. Overall, our method is the most computation-efficient to train \textit{from scratch}. As shown in Fig. \ref{fig:deep zero}, standard ZO training fails to converge well; \texttt{DeepZero} may eventually converge to a good solution, however, at the cost of $200\times$ more forward passes.

\begin{figure*}[t]
    \centering
    \includegraphics[width=\linewidth]{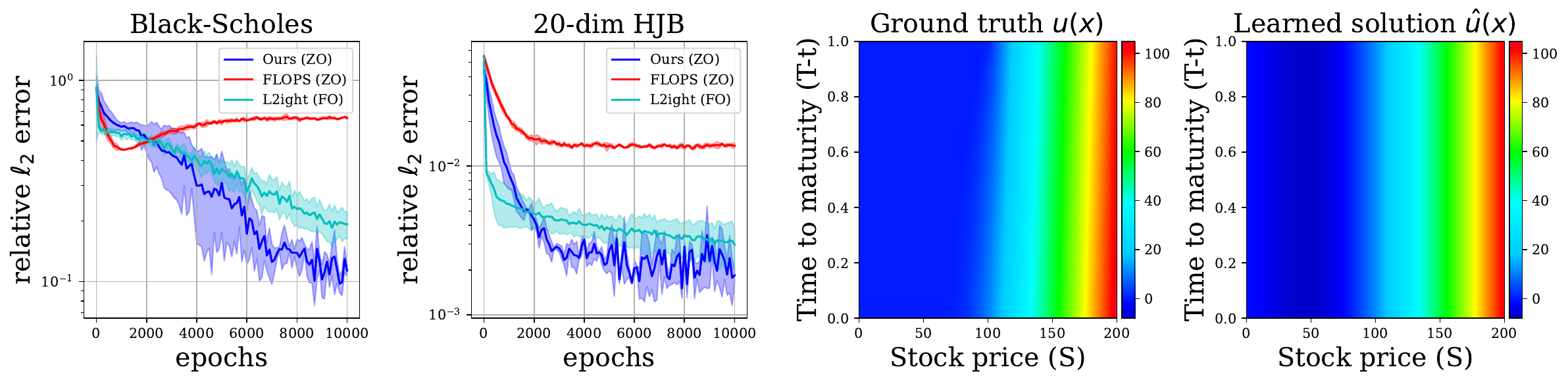}
     \vspace{-20pt}
    \caption{The first two subfigures show the relative $\ell_2$ error of Black-Scholes and 20-dim HJB equations learned by different ONN training methods. The last two subfigures show the ground truth $u(x)$, and the learned solution $\hat{u}(x)$ using our proposed method.}
    \label{fig:phase domain}
     \vspace{-10pt}
\end{figure*}


\subsection{On-chip Training Simulation} 
We further simulate on-chip \textit{phase-domain} training where the training parameters are MZI phases $\boldsymbol{\Phi}$ that parameterize the weight matrix $\mat{W}(\boldsymbol{\Phi})$ (TT-cores $\ten{G}(\boldsymbol{\Phi})$ in our proposed method). 

\paragraph{\bf ONN Simulation Set-ups.} 
We implemented the ONN simulation based on an open-source PyTorch-centric ONN library \texttt{TorchONN}. 
The linear projection in an ONN adopts blocking matrix multiplication, where the $M\times N$ weight matrix is partitioned into $P\times P$ blocks of size $k\times k$. Here $P=\lceil M/k \rceil, Q=\lceil N/k \rceil$. The weight matrix $\boldsymbol{W}$ is parameterized by MZI phases $\boldsymbol{\Phi}$ as $\boldsymbol{W}(\boldsymbol{\Phi})=\left\{\boldsymbol{W}_{p q}\left(\boldsymbol{\Phi}_{p q}\right)\right\}_{p=0, q=0}^{p=P-1, q=Q-1}$. Each block $\boldsymbol{W}_{pq}$ is parameterized as $\boldsymbol{W}_{p q}\left(\boldsymbol{\Phi}_{p q}\right)=\boldsymbol{U}_{p q}\left(\boldsymbol{\Phi}_{p q}^U\right) \boldsymbol{\Sigma}_{p q}\left(\boldsymbol{\Phi}_{p q}^S\right) \boldsymbol{V}_{p q}^*\left(\boldsymbol{\Phi}_{p q}^V\right)$. 
We provide more ONN Simulation details in Appendix \ref{appendix:ONN Simulation Settings}.
We follow~\cite{gu2021l2ight} to consider the following hardware-restricted objective $\bm{\Phi}^* = {\arg \min}_{\bm{\Phi}} \mathcal{L}(\bm{W}(\bm{\Omega} \bm{\Gamma} \mathcal{Q}(\bm{\Phi})+\bm{\Phi}_b))$, which jointly considers control resolution limit $\mathcal{Q}(\cdot)$, phase-shifter $\gamma$ coefficient drift $\bm{\Gamma}\sim \mathcal{N}(\gamma, \sigma^2_\gamma)$ caused by fabrication variations, thermal cross-talk between adjacent devices $\bm{\Omega}$, and phase bias due to manufacturing error $\bm{\Phi}_b \sim \mathcal{U}(0,2\pi)$.
We provide detailed set-ups in Appendix \ref{appendix:ONN Non-ideality}.

\paragraph{\bf Baseline Methods.} 
Table \ref{tab:photonics training} compares our method with existing on-chip BP-free ONN training methods, including \texttt{FLOPS}~\citep{GuDAC} and subspace training \texttt{L\textsuperscript{2}ight}~\citep{gu2021l2ight}. 
\texttt{FLOPs} \cite{GuDAC} is a ZO based method. We use zeroth-order gradient estimation to estimate the gradients of all MZI phases (i.e., $\boldsymbol{\Phi}_{pq}^U, \boldsymbol{\Phi}_{pq}^S, \boldsymbol{\Phi}_{pq}^V$)
\texttt{L\textsuperscript{2}ight} \cite{gu2021l2ight} is a subspace FO based method. Due to the intractable gradients for $\boldsymbol{\Phi}_{pq}^U$ and $\boldsymbol{\Phi}_{pq}^V$, only the MZI phase shifters in the diagonal matrix $\boldsymbol{\Phi}_{pq}^S$ are trainable. This restricts the training space (i.e., subspace training).

Note that existing methods do NOT support PINN training. We apply the same sparse-grid loss computation in all methods. 
We use the same number of ONN forward evaluations per step in different BP-free training methods for fair comparisons.

\paragraph{\bf Training Performance.} 
The first two subfigures in Fig. \ref{fig:phase domain} shows the relative $\ell_2$ error curves of different training protocols.
Real-size PINNs training are very high-dimensional optimization problems (18k MZIs for Black-Scholes, 280k MZIs for 20dim-HJB, and 73k MZIs for Burgers' and Darcy Flow) for ONN on-chip training.
\texttt{FLOPS} can only handle toy-size neural networks ($20\sim 30$ neurons per layer, $\sim 1k$ MZIs) and fail to converge well on real-size PINNs, thus is not capable of solving realistic PDEs due to the limited scalability. 
Subspace BP training method \texttt{L\textsuperscript{2}ight} enables on-chip FO training of ONN, however the trainable parameters are restricted to the diagonal matrix $\bm{\Sigma}(\bm{\Phi})$ while orthogonal matrices $\bm{U}(\bm{\Phi})$ and $\bm{V}(\bm{\Phi})$ are frozen at random initialization due to the intractable gradients. Such restricted learnable space hinders the degree of freedom for training PINNs from scratch. As a result, \texttt{L\textsuperscript{2}ight} only finds a roughly converged solution with a large relative $\ell_2$ error.
Our tensor-compressed BP-free training achieves the lowest relative $\ell_2$ error after on-chip training. 
We also visualize the learned solution $\hat{u}$ to examine the quality (the last two subfigures in Fig. \ref{fig:phase domain}).
The above results show that our method is the most scalable solution to enable real-size PINNs training, capable of solving realistic PDEs on photonic computing hardware. The on-chip phase-domain training results normally show some performance degradation compared with the numerical results of weight-domain training, due to the limited control resolution, device uncertainties, \textit{etc.}. More experiment results are available in Appendix \ref{appendix:Detailed Results of Phase-domain Training}.

\begin{table}[t]
  \centering
  \caption{Relative $\ell_2$ error in different photonic training methods. 
  }
  \resizebox{\linewidth}{!}{

    \begin{tabular}{c|ccc}
    \toprule
    \toprule
    Problem & {\texttt{FLOPS}} & {\texttt{L\textsuperscript{2}ight}} & Ours \\
          & \cite{GuDAC} & \cite{gu2021l2ight} &  \\
    \midrule
    Black-Scholes & 6.67E-01 & 2.03E-01 & \textbf{1.03E-01} \\
    20-dim HJB & 1.40E-02 & 4.09E-03 & \textbf{1.57E-03} \\
    Burgers & 4.47E-01 & 5.69E-01 & \textbf{2.68E-01} \\
    Darcy Flow & 4.76E-01 & 1.54E-01 & \textbf{9.10E-02} \\
    \bottomrule
    \bottomrule
    \end{tabular}%

  } 
  \label{tab:photonics training}%
 \vspace{-10pt}
\end{table}%

\begin{table}[t]
\centering
\caption{Implementation results a $128\times 128$ hidden layer in solving Black-Scholes equation. The latency means total on-chip training time. SM: space multiplexing. TM: time multiplexing.}
\resizebox{\linewidth}{!}{
    \begin{tabular}{cccc}
    \toprule
    \toprule
          & \# of MZIs & Footprint ($mm^2$) & Training time (s) \\
    \midrule
    ONN-SM & 16384 & 4206.08 (infeasible) & 1.74 \\
    Ours (w/ TONN-SM) & \textbf{384} & \textbf{102.72} & \textbf{1.64} \\
    \midrule
    ONN-TM & 64    & 19.52 & 52.27 \\
    Ours (w/ TONN-TM) & 64    & 19.52 & \textbf{9.80} \\
    \bottomrule
    \bottomrule
    \end{tabular}%
    
}
\label{tab:hardware-cost}%
\vspace{-10pt}
\end{table}

\begin{table*}[t]
\vspace{-10pt}
  \centering
  \caption{Footprint breakdown. All units are ${\rm mm}^2$.}
  \resizebox{0.7\linewidth}{!}{
    \begin{tabular}{ccccccc}
    \toprule
    \toprule
          & Laser & MRR Modulator & Tensor core & Photodetector & Cross-connect & Total \\
    \midrule
    ONN-SM & 25.6  & 1.28  & 4177.92 & 1.28  & /     & 4206.08 \\
    TONN-SM & 1.6   & 0.8   & 97.92 & 0.8   & 1.6   & 102.72 \\
    \midrule
    ONN-TM & 1.6   & 0.8   & 16.32 & 0.8   & /     & 19.52 \\
    TONN-TM & 1.6   & 0.8   & 16.32 & 0.8   & /     & 19.52 \\
    \bottomrule
    \bottomrule
    \end{tabular}%
    
    }
  \label{tab:footprint breakdown}%
\end{table*}%

\begin{table*}[t]
  \centering
  \caption{Latency breakdown. The results are based on simulation. ONN-SM and TONN-SM denote space-multiplexing implementation. ONN-TM and TONN-TM denote time-multiplexing implementation.}
  \resizebox{0.7\linewidth}{!}{
        \begin{tabular}{cccccc}
        \toprule
        \toprule
              & Cycles & Time per Inference (ns) & Time per epoch (ms) & Time to converge (s) & rel. $\ell_2$ error \\
        \midrule
        ONN-SM & 1     & 51.30 & 0.17  & 1.74  & 0.667 \\
        TONN-SM & 1     & 48.74 & 0.16  & \textbf{1.64}  & \textbf{0.103} \\
        \midrule
        ONN-TM & 32    & 1545.92 & 5.23  & 52.27 & 0.667 \\
        TONN-TM & 6     & 289.86 & 0.98  & \textbf{9.80}  & \textbf{0.103} \\
        \bottomrule
        \bottomrule
        \end{tabular}%
        
    }
  \label{tab:latency}%
\end{table*}%

\subsection{System Performance Evaluation}\label{system performance}
We use the Black-Scholes equation as an example to conduct pre-silicon system performance evaluation.
The performance for the accelerators based on ONNs and TONNs is evaluated and compared assuming the III-V-on-Si device platform~\cite{liang2022energy}. 
Table \ref{tab:hardware-cost} compares the device cost and footprint to implement a $128\times 128$ hidden layer and the total training time for solving the Black-Scholes equation. The model size is much larger than existing photonic training accelerators that only support $20\sim 30$ neurons per layer \cite{bandyopadhyay2022single, pai2023experimentally}. We compare our accelerator design with the conventional ONN design in both space multiplexing (SM) and time multiplexing (TM) designs. 
It is not practical for a single photonic chip to integrate a $128\times 128$ matrix due to the huge device sizes and the insurmountable optical loss.
In comparison, our method reduces the number of MZIs by $42.7\times$, which is the key to enabling whole-model integration (TONN-SM) with a reasonable footprint. 
The simulation results show that our photonic accelerator achieve ultra-high-speed PINN training (1.64-second) to solve the Black-Scholes equation. 
Detail breakdowns are as follow.

\subsubsection{Footprint:} 
The electronic part only performs lightweight scalar updates and signal routing (as described in Section \ref{label: photonic design}) thus consumes a minor part of the whole system. Following prior electronic-photonic accelerator design \cite{bandyopadhyay2022single, demirkiran2023electro}, we focus on the photonic footprint in our comparison. Table \ref{tab:footprint breakdown} summarizes the results. We computed the photonic footprint by Eq. \eqref{eq:footprint}  which includes the areas of hybrid silicon comb laser $A_{Laser}$, microring resonator (MRR) modulator arrays $A_{Modulator}$, photonic tensor cores, photodiodes $A_{PD}$, and electrical cross-connects $A_{Cross-connects}$. We attached the design parameters in Appendix \ref{appendix:Design Parameters for System Performance Evaluation}. 

\vspace{-10pt}
\begin{equation}
\begin{aligned}
A= & n_{\mathrm{MZI} \_ \text {mesh }} A_{\mathrm{MZI} \_ \text {mesh }}+ N A_{\text {Laser }}+2 N A_{\text {Modulator }} \\
& +2 N A_{\mathrm{PD}}+n_{\text {cross-connect }} A_{\text {Cross-connect }}
\end{aligned}
\label{eq:footprint}
\end{equation}


\subsubsection{Latency:} 

\paragraph{Latency per Inference.} 
The latency per inference is calculated by: 
\begin{equation}
    t_{\rm{inference}}=n_{\rm{cycle}}*(t_{\rm{DAC}}+t_{tuning}+t_{\rm{opt}}+t_{\rm{ADC}})
\end{equation}
where $t_{\rm{DAC}}$ is the DAC conversion delay ($\sim$24 ns), $t_{\rm{tuning}}$ is the metal-oxide-semiconductor capacitor (MOSCAP) phase shifter tuning delay ($\sim$0.1 ns), $t_{\rm{opt}}$ is the propagation latency of optical signal ($\sim$3.20 ns for ONN, $\sim$0.64 ns for TONN-SM, and $\sim$0.21 ns for TONN-TM), $t_{\rm{ADC}}$ is the ADC delay($\sim$24 ns). The TONN-TM uses 6 cycles for one inference, while ONN and TONN-SM only needs 1 cycle.
The latency per inference is estimated at 51.30 ns for ONN, 48.74 ns for TONN-SM, and 289.86 ns for TONN-TM.

\paragraph{Latency per Epoch.}
The latency per epoch is 
\begin{equation}
\small
    t_{\rm epoch} = (t_{\rm inference} \times N_{\rm point} \times N_{\rm loss} + t_{\rm tuning}) \times N_{\rm grads} +  t_{\rm DIG} \nonumber
\end{equation}
$t_{\rm{DIG}}$ is the digital computation overhead ($\sim$500 ns) for gradient accumulation and phase updates at the end of each epoch. 
New random perturbation samples could be sampled from environment in parallel with optical inference, so we didn't include this overhead.
We use $N_{\rm point}=130, N_{\rm loss}=13, N_{\rm grads}=2$.
The latency per epoch is estimated at 0.174 ms for ONN, 0.164 ms for TONN-SM, and 0.980 ms for TONN-TM.

\paragraph{Total Training Latency.} 
On average our BP-free training finds a good solution after 10000 epochs of update. The total training latency is estimated as 1.74 s for ONN, 1.64 s for TONN-SM, and 9.80 s for TONN-TM. Table \ref{tab:latency} summarizes the breakdown of training latency.

\paragraph{\bf Remark. } The model size that a photonic AI accelerator can handle is much smaller than its electronic counterparts, due to the larger sizes of photonic devices. It is worthnoting that our {\it training} accelerator can handle much larger neural network models than the state-of-the-art photonic {\it inference} accelerator~\citep{Ramey2020}.

\section{Conclusion and Future Work}
Motivated by the growing need for ultra-fast training accelerators to support physics-aware decision making and control in physical AI and digital twins, this paper has proposed a two-level BP-free approach to train real-size physics-informed neural networks (PINNs) on optical computing hardware. 
Specifically, our method integrates a sparse-grid Stein derivative estimator to avoid BP in loss evaluation and a tensor-compressed ZO optimization to avoid BP in model parameter update. The tensor-compressed ZO optimization can simultaneously reduce the ZO gradient variance and model parameters, thus scaling up optical training to real-size PINNs with hundreds of neurons per layer, which is much larger than existing photonic AI hardware. We have further presented an integrated photonic implementation and conducted comprehensive pre-silicon performance evaluations. Our approach has successfully solved various PDE benchmarks with the smallest relative error compared with existing photonic on-chip training protocols. 

Following the pre-silicon evaluation in this work, we plan to demonstrate the performance on a real electronic-photonic hardware prototype in the future.
On the algorithm side, future studies of variance reduction can help narrow the inevitable performance gap between ZO training and FO training. The BP-free training method presented in this work offers a generic routine towards hardware-efficient learning on physical edge. Our approach can be easily extended to other applications on photonic and other types of edge platform where the hardware cost to implement BP is not feasible. We refer the reader to Appendix \ref{apdx:Broader Impacts} for a discussion and additional results on the broader impacts.

\newpage

\bibliography{example_paper}
\bibliographystyle{mlsys2025}

\newpage
\appendix
\onecolumn

\section{ONN Basics}\label{appendix:ONN Basics}
\subsection{MZI-based ONN Architecture.}
We focus on the ONN~\citep{shen2017deep} architecture with singular value decomposition (SVD) to implement matrix-vector multiplication (MVM), i.e., $y=\bm{W}x=\bm{U\Sigma V^*}x$. The unitary matrices $\bm{U}$ and $\bm{V^*}$ are implemented by MZIs in Clements mesh~\citep{rectangleMZI}. 
The parametrization of $\bm{U}$ and $\bm{V^*}$ is given by $\bm{U}(\bm{\Phi}^U)=\bm{D}^U \prod_{i=k}^2 \prod_{j=1}^{i-1} \bm{R}_{i j}\left(\phi_{i j}^U\right), \bm{V}^*(\bm{\Phi}^V)=\bm{D}^V \prod_{i=k}^2 \prod_{j=1}^{i-1} \bm{R}_{i j}\left(\phi_{i j}^V\right)$, where $\bm{D}$ is a diagonal matrix, and each 2-dimensional rotator $\bm{R}_{ij}(\phi_{ij})$ can be implemented by a reconﬁgurable $2\times 2$ MZI containing one phase shifter ($\phi$) and two 50/50 splitters, which can produce interference of input light signals as follows:
\begin{equation}
\binom{y_1}{y_2}=\left(\begin{array}{cc}
\cos \phi & \sin \phi \\
-\sin \phi & \cos \phi
\end{array}\right)\binom{x_1}{x_2}
\end{equation}

The diagonal matrix $\boldsymbol{\Sigma}$ is implemented by on-chip attenuators, e.g., single-port MZIs, to perform signal scaling. The parameterization is given by $\boldsymbol{\Sigma}\left(\boldsymbol{\Phi}^S\right)=\max \left(\left|\boldsymbol{\Sigma}\right|\right) \operatorname{diag}\left(\cdots, \cos \phi_{i}^S, \cdots\right)$.
We denoted all programmable phases as $\bm{\Phi}$ and $\bm{W}$ is parameterized by $\bm{W}(\bm{\Phi})=\bm{U}(\bm{\Phi}^U) \bm{\Sigma}(\bm{\Phi}^S) \bm{V}^*(\bm{\Phi}^V)$. 

To implement a $N\times N$ matrix on ONN, $O(N^2)$ MZIs are required no matter the large ONN is implemented with a single large MZI mesh or multiple smaller MZI meshes. For a single MZI mesh implementation, the number of MZIs is $(\frac{N(N-1)}{2} + N + \frac{N(N-1)}{2})=N^2$. To implement with multiple smaller MZI meshes, saying implementing a $N\times N$ matrix by $\frac{N}{k}\times \frac{N}{k}$ blocks with size of with $k\times k$, the number of MZIs is $\frac{N}{k}\times \frac{N}{k}\times (\frac{k(k-1)}{2} + k + \frac{k(k-1)}{2})=N^2$.
Due to the high MZI cost and large MZI footprint, the space-multiplexing implementation of ONN is not realistic for large weight matrices. 

\subsection{Intractable Gradients of MZI Phases}
The analytical gradient w.r.t each MZI phases is given by:
\begin{equation}
\frac{\partial \mathcal{L}}{\partial \boldsymbol{R}_{i j}}=\left(\boldsymbol{D} \boldsymbol{R}_{n 1} \boldsymbol{R}_{n 2} \boldsymbol{R}_{n 3}\right)^T \nabla_y \mathcal{L} x^T\left(\cdots \boldsymbol{R}_{32} \boldsymbol{R}_{21} \boldsymbol{\Sigma} \boldsymbol{V}^*\right)^T
\end{equation}

\begin{equation}
\frac{\partial \mathcal{L}}{\partial \phi_{i j}}=\operatorname{Tr}\left(\left(\frac{\partial \mathcal{L}}{\partial \boldsymbol{R}_{i j}} \odot \frac{\partial \boldsymbol{R}_{i j}}{\partial \phi_{i j}}\right)\left(e_i+e_j\right)\left(e_i+e_j\right)^T\right)
\end{equation}

This analytical gradient is computationally-prohibitive, and requires detecting the whole optical field to read out all intermediate states $x$, which is not practical or scalable on integrated photonics chip.

\section{TONN Implementation Details}\label{appendix:TONN Implementation Details}
\subsection{ Additional details on TONN-SM Architecture}\label{apdx: Additional details on TONN-SM Architecture}

\begin{figure}[t]
\centering

\includegraphics[width=0.9\textwidth]{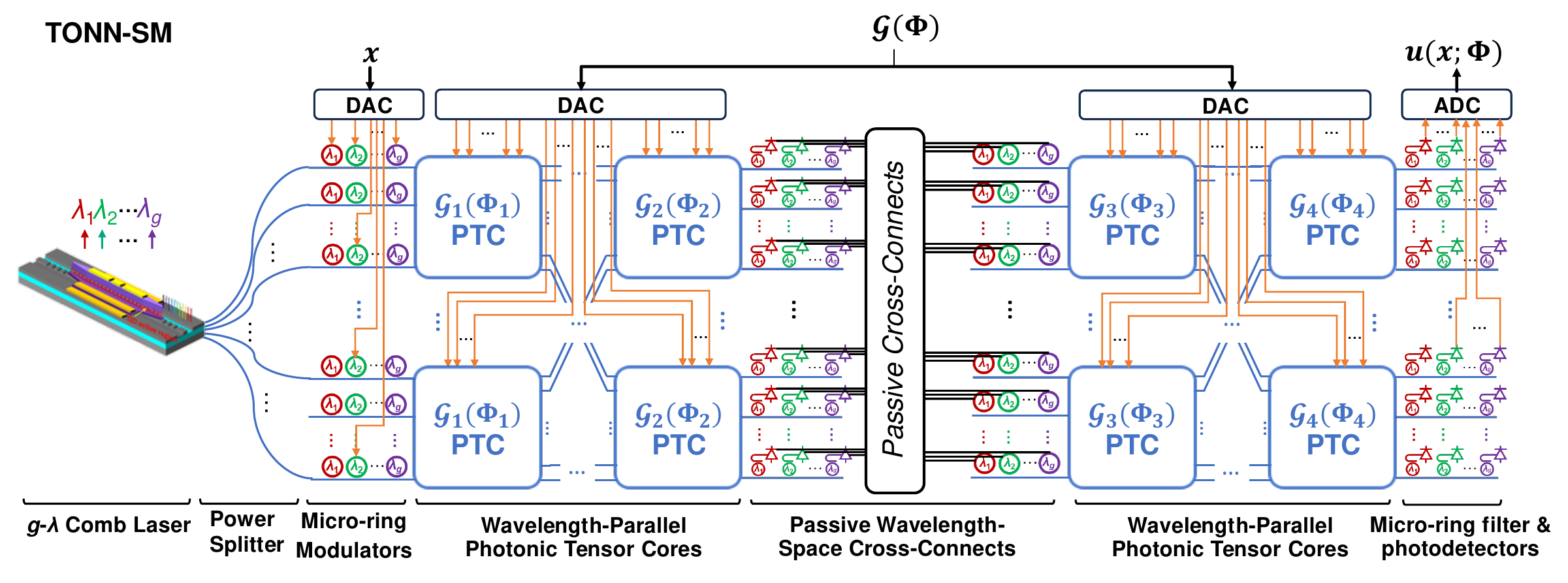}
\caption{\label{fig:apdx_TONN_1} (The same as Figure \ref{fig:TONN} (c)) TONN-SM architecture. PTC: photonic tensor core, DAC: digital-analog converter, ADC: analog-digital converter.}

\end{figure}
    
{
 In TONN-SM, the input data $\textbf{x}\in \mathbb{R}^{N}$, is folded to a d-way tensor $\ten{X} \in \mathbb{R}^{N_d\times \cdots \times N_1}$. The indices of the input tensor is then represented by $g$ wavelength division multiplexing (WDM) channels at $N/g$ inputs of the tensor cores, where $g = N_{d/2}\times \ldots \times N_1$. The light source is provided by a $g$-wavelength comb laser and power splitters. The splitted WDM light is modulated by $g$-wavelength optical modulator arrays, then multiplied by each of the photonic tensor core layers, and finally detected by $g$-wavelength WDM microring add-drop filter and detector arrays. The photonic tensor core layer $k$ ($k=d, \ldots, 1, k\neq d/2+1$) consists of $h_k$ number of $R_{k-1}M_k\times N_kR_k$ MZI meshes (tensor cores) and an optical passive cross-connect to switch indices of $M_k$ and $N_{k-1}$. Here, $h_k = M_d\ldots M_{k+1}N_{k-1}\ldots N_{d/2+1}$ for $d/2 \textless k \textless d$ or $M_{d/2}\ldots M_{k+1}N_{k-1}\ldots N_1$ for $k\leq d/2$. For TT-core $d/2+1$, the optical passive cross-connect is replaced by a passive wavelength-space cross-connect to switch the indices between the wavelength domain ($N_{d/2}, \ldots, N_1$) and the space domain ($M_d, \ldots, M_{d/2+1}$).
}

\subsection{Bit Accuracy}
In this paper, we assume a weight stationary (WS) scheme, where the weight matrices are programmed into the phase shifters in the MZI mesh, and the input vectors are encoded in the high-speed (10 GHz) optical signals. In each training iteration, the same weights (phases) are multiplied with batched (e.g., 1000) input data. As a result, the update rate of the phase shifters is 10 GHz/1000 = 10 MHz. In a system-level study of MZI-mesh-based photonic AI accelerators, a 12-bit DAC is enough to support the 8-bit accuracy of the weights \cite{demirkiran2023electro}. Considering that a 12-bit DAC with a 10 MHz sampling rate is very mature \cite{TIADC}, assuming 8-bit weights (phases) in our setting is reasonable.

The minimum optical SNR at the output of the MZI mesh is $SNR = 2^{b_{out}}$, where $b_{out}$ is the required bit accuracy of the output of the matrix multiplication. The optical SNR can be improved by increasing the input laser power, reducing the optical insertion loss, increasing the optical gain, and increasing the sensitivity of the photodiodes. For instance, the platform in \cite{liang2022energy} can provide lasers with high wall-plug efficiency, optical modulators, and MZIs with low insertion loss, on-chip optical gain, and quantum dot avalanche photodiodes with low sensitivity. Furthermore, the tensor decomposition in our work reduces the number of cascaded stages of MZIs, significantly reducing the insertion loss induced by cascaded MZIs.

\subsection{Interconnection between Digital Control System ad TONN}
The digital control system is implemented via electronic-photonic co-integration that contains an FPGA or ASIC for controlling and digital calculations required by BP-free training, digital electronic memory (e.g., DRAM) for weight and data storage and buffering, and ADC/DACs for converting the digital data to the tuning voltages of the modulators and phase shifters. As a result, no additional optical devices are required other than the TONN inference accelerator we introduced in the paper. The noise induced by the digital control system is decided by the bit accuracy of the ADCs and DACs. Regarding synchronization between the digital system and TONN. In the WS scheme, weight buffers are used, which means that the weights for the next set of matrix multiplication are loaded into the weight buffer, while the MZI mesh performs matrix multiplication with the current weight values. The latency is limited by the tuning mechanism of the phase shifters. In our case, the tuning mechanism is the III-V-on-silicon metal-oxide-semiconductor capacitor (MOSCAP) \citep{liang2022energy}, which has a modulation speed of tens of GHz.

\section{Experiment Settings}\label{appendix:Training Set-ups}
\subsection{PDE details} \label{appendix:PDE details}
\paragraph{1-dim Black-Scholes Equation.} 
We examine the Black-Scholes equation for option price dynamics:
\begin{equation}
\begin{aligned}
&\partial_t u + \frac{1}{2}\sigma^2 x^2 \partial_{xx} u + rx\partial_x u - ru = 0, \quad x \in [0,200],~~t \in [0,T],\\
&u(x,T) = \max(x-K,0), \quad x \in [0,200],  \\
&u(0,t) = 0, \quad u(200,t) = 200 - Ke^{-r(T-t)}, \quad t \in [0,T],
\end{aligned}
\end{equation}
where $u(x,t)$ is the option price, $x$ is the stock price, $\sigma=0.2$ is volatility, $r=0.05$ is risk-free rate, $K=100$ is strike price, and $T=1$ is expiration time. The analytical solution is:
\begin{equation}
u(x,t) = xN(d_1) - Ke^{-r(T-t)}N(d_2),
\end{equation} with $d_1$ and $d_2$ defined as:
\begin{equation}
\begin{aligned}
&d_1 = \frac{\ln(x/K) + (r + \sigma^2/2)(T-t)}{\sigma\sqrt{T-t}}, \\
&d_2 = d_1 - \sigma\sqrt{T-t},
\end{aligned}
\end{equation}
where $N(\cdot)$ is the cumulative distribution function of the standard normal distribution. The base neural network is a 3-layer MLP with 128 neurons and \texttt{tanh} activation in each hidden layer. In tensor-train (TT) compressed training, the input layer ($2\times 128$) and the output layer ($128\times 1$) are left as-is, while we fold the hidden layer as size $4\times 4\times 8\times 8\times 4\times 4$. We preset the TT-ranks as [1,$r$,$r$,1], where $r$ controls the compression ratio.

\paragraph{20-dim HJB Equation.}
We consider the following 20-dim HJB PDE for high-dimensional optimal control:
\begin{equation}
\begin{aligned}
&\partial_t u(\bm{x}, t)+\Delta u(\bm{x}, t)-0.05 \left\|\nabla_{\bm{x}}u(\bm{x}, t)\right\|_{2}^{2}=-2, \\
&u(\bm{x}, 1)=\left\|\bm{x}\right\|_{1}, \quad \bm{x} \in [0,1]^{20}, ~~t \in[0, 1].
\end{aligned}
\end{equation}
Here $\left\|\cdot\right\|_{p}$ denotes an $\ell_p$ norm. The exact solution is $u(\bm{x},t)=\left\|\bm{x}\right\|_{1}+1-t$. The base network is a 3-layer MLP with 512 neurons and \texttt{sine} activation in each hidden layer. For TT compression, we fold the input layer and hidden layers as size $1\times 1\times 3\times 7\times 8\times 4\times 4\times 4$ and $4\times 4\times 4\times 8\times 8\times 4\times 4\times 4$, respectively,with TT-ranks [1,$r$,$r$,$r$,1]. The output layer ($512\times 1$) is left as-is.

{

\textbf{1-dim Burgers’ Equation \cite{hao2023pinnacle}:}
\begin{equation}
    \partial_t u + u \partial_x u = \nu  \partial_{xx} u, \quad (x, t) \in [-1, 1] \times [0, 1],
\end{equation}
where the viscosity $\nu = \frac{0.01}{\pi}$. The initial and boundary conditions are:
\begin{align}
    u(x, 0) &= -\sin(\pi x), \quad x \in [-1, 1], \\
    u(-1, t) &= u(1, t) = 0, \quad t \in [0, 1].
\end{align}

\textbf{2-dim Darcy Flow \cite{li2020fourier}:}
\begin{align}
\nabla \cdot(k(\mathbf{x}) \nabla u(\mathbf{x}))=f(\mathbf{x}), \quad \mathbf{x} \in \Omega,\\
u(\mathbf{x}) = 0, \quad \mathbf{x} \in \partial \Omega,
\end{align}
where $k(\mathbf{x})$ is the permeability field, $u(\mathbf{x})$ is the pressure, and $f(\mathbf{x})$ is the forcing function. We define $\Omega = [0, 1]^2$, set $f(\mathbf{x}) = 1$, and use a piecewise constant function for $k(\mathbf{x})$ as shown in Fig.~\ref{fig:darcy kx}.

\begin{figure}[H]
    \centering
    \includegraphics[width=0.5\linewidth]{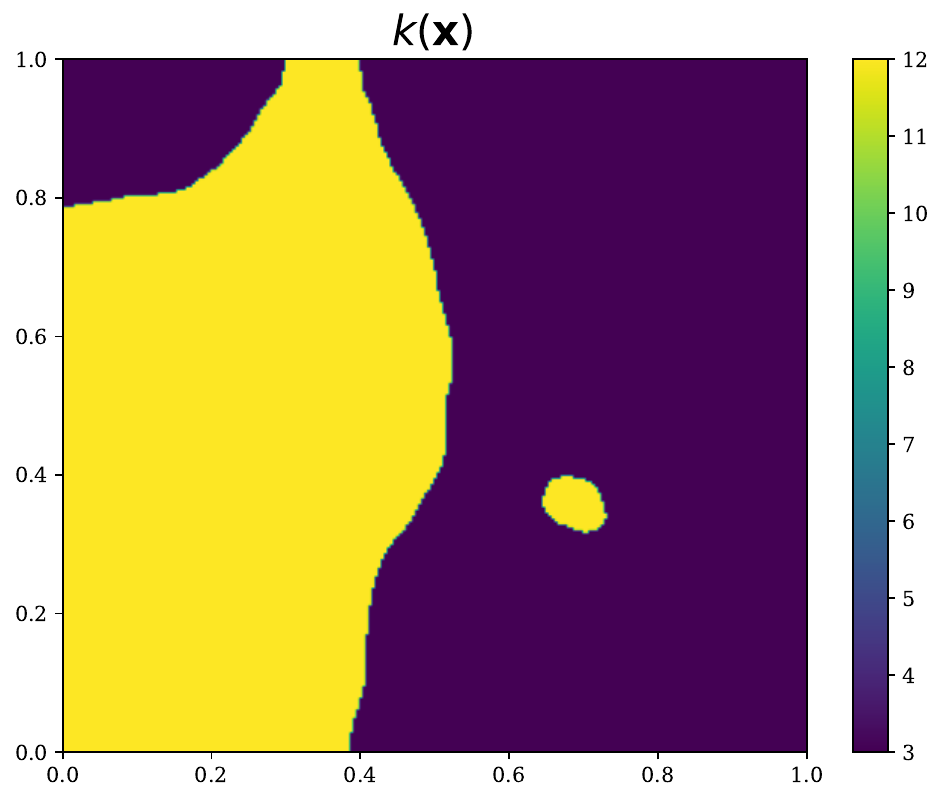}
    \caption{ Permeability field in the Darcy flow problem.}
    \label{fig:darcy kx}
\end{figure}
}

For both 1-dim Burgers' equation and 2-dim Darcy Flow, our baseline model aligns with the state-of-the-art PINN benchmark from \cite{hao2023pinnacle}. It comprises a fully connected neural network with five hidden layers containing 100 neurons, totaling 30,701 trainable parameters. The dimension of our tensor-compressed training is reduced to 1,241 by folding the weight matrices in hidden layers as size $4\times 5\times 5\times 5\times 5\times 4$ and decomposing it with a TT-rank $(1, 2, 2, 1)$. We trained the models for 40,000 iterations on the 1-dim Burgers' equation and for 20,000 iterations on the 2-dim Darcy flow. All other training configurations were kept consistent with our main experimental setups.

\subsection{Loss Evaluation Set-ups.}\label{appendix:Loss Evaluation Set-ups} We compare three methods for computing derivatives in the loss function \eqref{PINNs loss}: 1) automatic differentiation (\textbf{AD}) as a golden reference, 2) Monte Carlo-based Stein Estimator (\textbf{SE})~\cite{he2023learning}, and 3) our sparse-grid (\textbf{SG}) method. 
For Black-Scholes, we approximate the solution $u_{\bm{\theta}}$ using a neural network $f_{\bm{\theta}}(\bm{x},t)$, which can be either the base network or its TT-compressed version. In the \textbf{AD} approach, $u_{\bm{\theta}}(\bm{x},t)=f_{\bm{\theta}}(\bm{x},t)$, while for \textbf{SE} and \textbf{SG}, $u_{\bm{\theta}}(\bm{x},t)=\mathbb{E}_{(\bm{\delta_{\bm{x}}},\delta_{t}) \sim \mathcal{N}\left(\bm{0}, \sigma^{2} \bm{I}\right)}f_{\bm{\theta}}(\bm{x}+\bm{\delta_{\bm{x}}},t+\delta_{t})$. We set the noise level $\sigma$ to 1e-3 in \textbf{SE} and \textbf{SG}, using 2048 samples in \textbf{SE} and 13 samples in \textbf{SG} with a level-3 sparse Gaussian quadrature rule to approximate the expectations \eqref{gaussian smoothed model} and \eqref{stein derivative estimator}. For HJB, we employ a transformed neural network $f_{\bm{\theta}}^{'}(\bm{x},t) = (1-t)f_{\bm{\theta}}(\bm{x},t) + \left\|\bm{x}\right\|_{1}$, where $f_{\bm{\theta}}(\bm{x},t)$ is the base or TT-compressed network. The solution approximation follows the same pattern as in the Black-Scholes case. Here the transformed network is designed to ensure that our approximated solution either exactly satisfies (\textbf{AD}) or closely adheres to the terminal condition (\textbf{SE}, \textbf{SG}), allowing us to focus solely on minimizing the HJB residual during training. We set the noise level $\sigma$ to 0.1 in \textbf{SE} and \textbf{SG}, using 1024 samples in \textbf{SE} and 925 samples in \textbf{SG} with a level-3 sparse Gaussian quadrature rule.

\subsection{Training Set-ups.} We implemented all methods in PyTorch, utilizing an NVIDIA GTX 2080Ti GPU and an Intel(R) Xeon(R) Gold 5218 CPU @ 2.30GHz.

\subsection{Data Sampling.} For Black-Scholes, we uniformly sample 100 random residual points, 10 initial points, and 10 boundary points on each boundary per epoch to evaluate the PDE loss \eqref{PINNs loss}. For HJB, we select 100 random residual points per epoch. The model architecture for the HJB equation incorporates the terminal condition, eliminating the need for an additional terminal loss term. For Burgers', we uniformly sample 1200 random residual points, 100 initial points, and 100 boundary points on each boundary per epoch to evaluate the PDE loss \eqref{PINNs loss}. For Darcy flow, we sample the residual points on a fixed 241 x 241 uniform grid and encode hard boundary constraints in the model architecture.

\section{Weight-domain Training}\label{appendix:Detailed Results of Weight-domain Training}

\begin{figure}[H]
    \vspace{-10pt}
    \centering
    \includegraphics[width=0.8\linewidth]{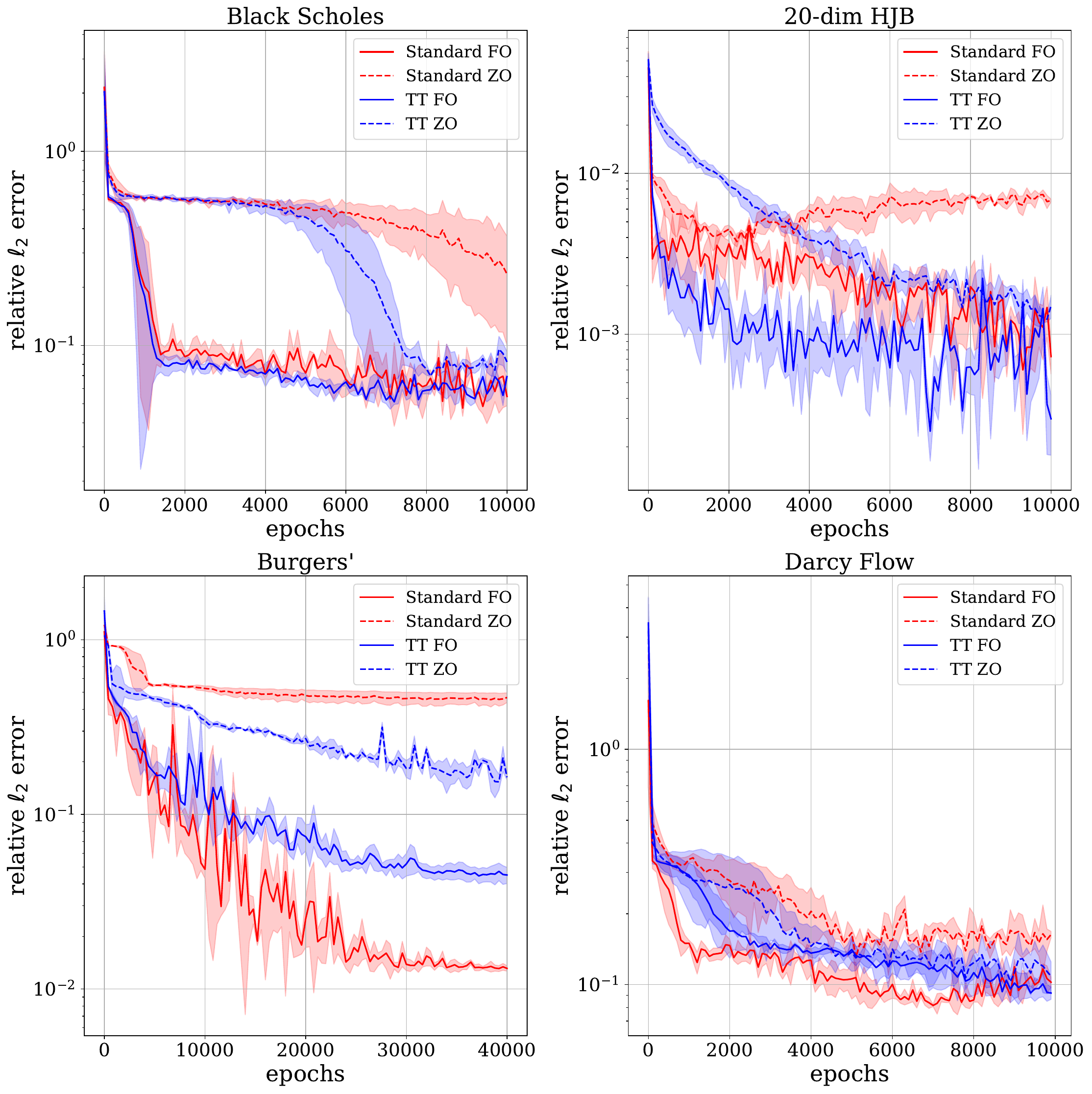}
        \caption{Relative $\ell_2$ error curves of weight domain training for Black-Scholes equation (left) and 20-dim HJB equation (right), respectively. The value at each step is averaged across three runs, and the shade indicates the standard deviation. 
    }
    \vspace{-20pt}
\label{fig:apdx_weight}
\end{figure}

{
 In this section, we provide the training curves of weight-domain training in Fig. \ref{fig:apdx_weight}. The curves denote averaged relative $\ell_2$ error over three independent experiments and shades denote the corresponding standard deviations. 
 We also provide the extended results of Table \ref{tab:loss computation} and \ref{tab:weight training}. Each relative $\ell_2$ error takes the form $\text{mean} \pm \text{std}$, where $\text{mean}$ denotes the averaged result over three independent experiments, and $\text{std}$ denotes the corresponding standard deviation. 
}

\begin{table}[H]
  \centering
  \caption{Relative $\ell_2$ error of FO training using
different loss computation methods. We report the averaged results and standard deviations across three runs.}
    \begin{tabular}{c|ccc}
    \toprule
    \toprule
    Problem & AD    & SE    & SG (ours) \\
    \midrule
    Black-Scholes & (5.35$\pm$0.13)E-02 & (5.41$\pm$0.09)E-02 & \boldmath{}\textbf{(5.28$\pm$0.05)E-02}\unboldmath{} \\
    20-dim HJB & (1.99$\pm$0.15)E-03 & (1.52$\pm$0.14)E-03 & \boldmath{}\textbf{(8.16$\pm$1.24)E-04}\unboldmath{} \\
    Burgers & (1.37$\pm$0.04)E-02 & (1.98$\pm$0.15)E-02 & \boldmath{}\textbf{(1.31$\pm$0.05)E-02}\unboldmath{} \\
    Darcy Flow & (7.57$\pm$0.28)E-02 & (7.85$\pm$0.40)E-01 & \boldmath{}\textbf{(7.47$\pm$0.41)E-02}\unboldmath{} \\
    \bottomrule
    \bottomrule
    \end{tabular}%

  \label{tab:addlabel}%
\end{table}%

\begin{table}[H]
  \centering
  \caption{Relative $\ell_2$ error achieved using different training methods. We report the averaged results and standard deviations across three runs.}
    \begin{tabular}{c|cc|cc}
    \toprule
    \toprule
    Problem & \multicolumn{2}{c|}{FO Training} & \multicolumn{2}{c}{ZO Training} \\
    \cmidrule{2-5}      & Standard & TT    & Standard & TT \\
    \midrule
    Black-Scholes & (5.28$\pm$0.05)E-02 & \boldmath{}\textbf{(5.97$\pm$0.01)E-02}\unboldmath{} & (3.91$\pm$0.05)E-01 & \boldmath{}\textbf{(8.30$\pm$0.08)E-02}\unboldmath{} \\
    20-dim HJB & (8.16$\pm$1.24)E-04 & \boldmath{}\textbf{(2.05$\pm$0.39)E-04}\unboldmath{} & (6.86$\pm$0.27)E-03 & \boldmath{}\textbf{(1.54$\pm$0.35)E-03}\unboldmath{} \\
    Burgers & (1.31$\pm$0.05)E-02 & \boldmath{}\textbf{(4.49$\pm$0.58)E-02}\unboldmath{} & (4.41$\pm$0.09)E-01 & \boldmath{}\textbf{(1.63$\pm$0.25)E-02}\unboldmath{} \\
    Darcy Flow & \boldmath{}\textbf{(7.47$\pm$0.41)E-02}\unboldmath{} & \boldmath{}\textbf{(8.77$\pm$0.11)E-02}\unboldmath{} & (1.34$\pm$0.18)E-01 & \boldmath{}\textbf{(9.05$\pm$0.29)E-02}\unboldmath{} \\
    \bottomrule
    \bottomrule
    \end{tabular}%
    
  \label{tab:addlabel}%
\end{table}%

\section{Ablation Studies}\label{appendix:Ablation Studies}
We conduct ablation studies on the hyperparameters. For simplicity we use a constant learning rate schedule with the Adam optimizer, as we empirically found that tuning the learning rate scheduler does not help or harm the training. The chosen hyperparameters are highlighted in \textbf{bold} in the results tables.

\subsection{Tensor-train (TT) Ranks}
{
TT-rank determination is a trade-off between model compression ratio and model expressivity. The TT-ranks can be empirically determined, or adaptively determined by automatic rank determination algorithms \cite{hawkins2021bayesian, yang2024comera}. To validate our tensor-train (TT) rank choice, we add an ablation study on different TT ranks. The results are provided in Table \ref{tab:abla_ttrank} below. We tested tensor-train compressed training with different TT-ranks on solving 20-dim HJB equations. The model setups are the same as illustrated in Appendix A.2. We fold the input layer and hidden layers as size $1\times 1\times 3\times 7\times 8\times 4\times 4\times 4$ and $4\times 4\times 4\times 8\times 8\times 4\times 4\times 4$, respectively, with TT-ranks [1,$r$,$r$,$r$,1]. We use automatic differentiation for loss evaluation and first-order (FO) gradient descent to update model parameters. Other training setups are the same as illustrated in Appendix A.3.  The results reveal that models with larger TT-ranks have better model expressivity and achieve smaller relative $\ell_2$ error. However, increasing TT-ranks increases the hardware complexity (e.g., number of MZIs) of photonics implementation as it increases the number of parameters. Therefore, we chose a small TT-rank as 2, which provides enough expressivity to solve the PDE equations, while maintaining a small model size.
}

\begin{table}[H]
  \centering
  \caption{Ablation study on tensor-train (TT) ranks when training the TT compressed model on solving 20-dim HJB equations. We report the average error and the standard deviation across three runs.}
    \begin{tabular}{ccccc}
    \toprule
    \toprule
    TT-rank & \textbf{2}     & 4     & 6     & 8 \\
    \midrule
    Params & 1,929 & 2,705 & 3,865 & 5,409 \\
    rel. $\ell_2$ error & (3.17$\pm$1.16)E-04 & (2.45$\pm$0.82)E-04 & (4.00$\pm$3.69)E-05 & (3.02$\pm$3.16)E-05 \\
    \bottomrule
    \bottomrule
    \end{tabular}%
  \label{tab:abla_ttrank}%
\end{table}%

\subsection{Hidden layer width of baseline MLP model}
{
We also performed an ablation study on the hidden layer width of the baseline MLP model. We trained 3-layer MLPs with different hidden layer widths to solve the 20-dim HJB equation. We use automatic differentiation for loss evaluation and first-order (FO) gradient descent to update model parameters. Other training setups are the same as illustrated in Appendix A.3. The results are shown in Table \ref{tab:abla_hidden}. The MLP model with a smaller hidden layer width leads to larger testing errors. This indicates that a large hidden layer is favored to ensure enough model expressivity. The MLP model used in our submission does not have an overfitting problem. 
}

\begin{table}[H]
  \centering
  \caption{Ablation study on hidden layer size of baseline 3-layer MLP model when learning 20-dim HJB equation. We report the average error and the standard deviation across three runs.}

    \resizebox{\linewidth}{!}{
    \begin{tabular}{cccccc}
    \toprule
    \toprule
    Hidden layer size & \textbf{512}   & 256   & 128   & 64    & 32 \\
    \midrule
    Params & 274,433 & 71,681 & 19,457 & 5,633 & 1,793 \\
    rel. $\ell_2$ error & (2.72$\pm$0.23)E-03 & (4.31$\pm$0.19)E-03 & (7.51$\pm$0.36)E-03 & (8.15$\pm$0.67)E-03 & (9.25$\pm$0.27)E-03 \\
    \bottomrule
    \bottomrule
    \end{tabular}%
    }
  \label{tab:abla_hidden}%
\end{table}%

\subsection{Finer 2D/3D Mesh Grids}

We have added an ablation study focused on our method’s performance under different mesh grid sizes of input samples. We applied our proposed fully backpropagation-free PINNs training method, which combines sparse-grid loss computation and tensor-train compressed zeroth-order training, to solve the Black-Scholes equation. Other experimental setups follow those used in our main experiments.

As shown in Table~\ref{tab:r9}, allowing a larger input sample mesh grid can slightly reduce the converged relative $\ell_2$ error. However, considering the computation budget, we chose mesh grid size as $100\times 100$.

\begin{table}[h]
\centering
\begin{tabular}{lccc}
\toprule
Grid Size & \textbf{$100 \times 100$} & $1000 \times 1000$ & $10000 \times 10000$ \\
\midrule
rel.\ $\ell_2$ error & 8.30E-02 & 7.38E-02 & 7.19E-02 \\
\bottomrule
\end{tabular}
\caption{Effect of mesh grid resolution on solution accuracy.}
\label{tab:r9}
\end{table}

\subsection{Sparse-Grid Derivative Estimator}

\subsubsection{Solving the Black-Scholes equation}:

The results are shown in Table ~\ref{tab:r1}, ~\ref{tab:r3}, \ref{tab:r4}. In summary:
\begin{itemize}
    \item Reducing the number of samples of Monte Carlo-based Stein estimator from 2048 to 64 harms the training convergence. 
    \item Our proposed sparse-grid Stein derivative estimator uses much fewer samples than Monte Carlo-based Stein estimator, and achieves smaller test error on convergence. Using level 3 with $n_L^* = 13$ can already achieve low error.
    \item $\sigma$ is robust within a wide range of selection.
\end{itemize}

\begin{table}[h]
\centering
\begin{tabular}{lccc}
\toprule
Number of Samples & 64 & 512 & \textbf{2048} \\
\midrule
rel.\ $\ell_2$ error & 1.96E-01 & 5.89E-02 & 5.41E-02 \\
\bottomrule
\end{tabular}
\caption{Effect of Monte Carlo sample size on Black-Scholes solution error.}
\label{tab:r1}
\end{table}

\begin{table}[h]
\centering
\begin{tabular}{lccc}
\toprule
Level & 2 & \textbf{3} & 4 \\
\midrule
Number of Samples & 5 & \textbf{13} & 29 \\
rel.\ $\ell_2$ error & 6.41E-02 & 5.28E-02 & 5.20E-02 \\
\bottomrule
\end{tabular}
\caption{Ablation study on number of samples $n_L^*$.}
\label{tab:r3}
\end{table}

\begin{table}[!h]
\centering
\begin{tabular}{lcccc}
\toprule
$\sigma$ & 0.1 & 0.01 & \textbf{0.001} & 0.0001 \\
\midrule
rel.\ $\ell_2$ error & 6.46E-02 & 6.40E-02 & 5.28E-02 & 5.63E-02 \\
\bottomrule
\end{tabular}
\caption{Ablation study on $\sigma$.}
\label{tab:r4}
\end{table}

\subsubsection{Convergence Analysis for Synthetic Functions}:

We evaluate the Laplacian of a synthetic function $u(x, y) = \mathbb{E}_{\delta \sim \mathcal{N}(\mathbf{0}, \sigma^2 \mathbf{I})} [f(x + \delta_x, y + \delta_y)]$ with $f(x, y) = e^{-\frac{\sigma^2}{2}} e^{-x} \sin(y)$. This gives the analytical solution $u(x, y) = e^{-x} \sin(y)$ and $\Delta u = 0$.

We set $\sigma = 0.1$ and evaluated the Laplacian on a $100 \times 100$ grid over $[0,1]^2$. The $\ell_2$ errors and number of function queries for both estimators are shown in Tables~\ref{tab:r7} and~\ref{tab:r8}.

\begin{table}[h]
\centering
\begin{tabular}{lcc}
\toprule
Number of Samples & $\ell_2$ Error & \# Function Queries \\
\midrule
1,024 & 10.7437 & 4096 \\
2,048 & 7.6796 & 8192 \\
4,096 & 5.4245 & 16384 \\
8,192 & 3.8732 & 32768 \\
16,384 & 2.7016 & 65536 \\
\bottomrule
\end{tabular}
\caption{Monte Carlo estimator performance for Laplacian computation.}
\label{tab:r7}
\end{table}

\begin{table}[h]
\centering
\begin{tabular}{lcccc}
\toprule
Accuracy Level & \# Samples & $\ell_2$ Error & \# Function Queries \\
\midrule
3 & 13 & 0.1142 & 52 \\
4 & 29 & 2.8217e-07 & 116 \\
5 & 53 & 4.0797e-08 & 212 \\
6 & 89 & 8.4356e-13 & 356 \\
7 & 137 & 8.7094e-13 & 548 \\
\bottomrule
\end{tabular}
\caption{Sparse-grid estimator performance for Laplacian computation.}
\label{tab:r8}
\end{table}

\textbf{Conclusion:} The sparse-grid Stein estimator achieves significantly lower errors with substantially fewer computational resources than the Monte Carlo estimator. For instance, at accuracy level 4, the sparse-grid method achieves an $\ell_2$ error of $2.82 \times 10^{-7}$ using only 29 samples, whereas the Monte Carlo method with 16,384 samples yields an error of 2.7016.

\subsection{Dimension-Reduced ZO Optimization via Tensor-Train Decomposition}

\begin{itemize}
    \item In Table~\ref{tab:r5}, using a relatively large $\mu$ (0.1) may not converge well. With a smaller $\mu$ ($\leq 0.01$), the final converged error is robust within a wide range of $\mu$ selection.
    \item In Table~\ref{tab:r6}, comparing at the same computation budget (10,000 forwards), $N=1$ achieved the smallest relative $\ell_2$ error.
\end{itemize}

\begin{table}[h]
\centering
\begin{tabular}{lcccc}
\toprule
$\mu$ & 0.1 & \textbf{0.01} & 0.001 & 0.0001 \\
\midrule
rel.\ $\ell_2$ error & 5.61E-01 & 8.47E-02 & 8.62E-02 & 8.83E-02 \\
\bottomrule
\end{tabular}
\caption{Ablation study of smoothing factor $\mu$.}
\label{tab:r5}
\end{table}

\begin{table}[!h]
\centering
\begin{tabular}{lcccc}
\toprule
N & \textbf{1} & 10 & 50 & 100 \\
\midrule
10,000 forwards & 8.47E-02 & 3.66E-01 & 5.25E-01 & 5.42E-01 \\
\bottomrule
\end{tabular}
\caption{Ablation study of query number $N$ under fixed budget.}
\label{tab:r6}
\end{table}

\section{Phase-domain Training}

\subsection{ONN Simulation Settings} \label{appendix:ONN Simulation Settings}
{
We apply the same setups as that in \texttt{L\textsuperscript{2}ight} \cite{gu2021l2ight} to implement uncompressed ONNs in baseline methods \texttt{FLOPS} \cite{GuDAC} and \texttt{L\textsuperscript{2}ight} \cite{gu2021l2ight}. The linear projection in an ONN adopts blocking matrix multiplication, where the $M\times N$ weight matrix is partitioned into $P\times P$ blocks of size $k\times k$. Here $P=\lceil M/k \rceil, Q=\lceil N/k \rceil$. Implementing ONNs with smaller MZI blocks is more practical and robust, and provides enough trainable parameters ($N^2/k$ singular values) for first-order based method \texttt{L\textsuperscript{2}ight}. Following the analysis provided in \cite{gu2021l2ight}, we select $k=8$ for practical consideration. 

The weight matrix $\boldsymbol{W}$ is parameterized by MZI phases $\boldsymbol{\Phi}$ as $\boldsymbol{W}(\boldsymbol{\Phi})=\left\{\boldsymbol{W}_{p q}\left(\boldsymbol{\Phi}_{p q}\right)\right\}_{p=0, q=0}^{p=P-1, q=Q-1}$. Each block $\boldsymbol{W}_{pq}$ is parameterized as $\boldsymbol{W}_{p q}\left(\boldsymbol{\Phi}_{p q}\right)=\boldsymbol{U}_{p q}\left(\boldsymbol{\Phi}_{p q}^U\right) \boldsymbol{\Sigma}_{p q}\left(\boldsymbol{\Phi}_{p q}^S\right) \boldsymbol{V}_{p q}^*\left(\boldsymbol{\Phi}_{p q}^V\right)$. 

\texttt{FLOPs} \cite{GuDAC} is a ZO based method. We use zeroth-order gradient estimation to estimate the gradients of all MZI phases (i.e., $\boldsymbol{\Phi}_{pq}^U, \boldsymbol{\Phi}_{pq}^S, \boldsymbol{\Phi}_{pq}^V)$

\texttt{L\textsuperscript{2}ight} \cite{gu2021l2ight} is a subspace FO based method. Due to the intractable gradients for $\boldsymbol{\Phi}_{pq}^U$ and $\boldsymbol{\Phi}_{pq}^V$, only the MZI phase shifters in the diagonal matrix $\boldsymbol{\Phi}_{pq}^S$ are trainable. This restricts the training space (i.e., subspace training).
}

\subsection{ONN Non-ideality}\label{appendix:ONN Non-ideality}
We follow~\cite{gu2021l2ight} to consider the following hardware-restricted objective $\bm{\Phi}^* = {\arg \min}_{\bm{\Phi}} \mathcal{L}(\bm{W}(\bm{\Omega} \bm{\Gamma} \mathcal{Q}(\bm{\Phi})+\bm{\Phi}_b))$, which jointly considers various ONN non-ideality including control resolution limit $\mathcal{Q}(\cdot)$, phase-shifter $\gamma$ coefficient drift $\bm{\Gamma}\sim \mathcal{N}(\gamma, \sigma^2_\gamma)$ caused by fabrication variations, thermal cross-talk between adjacent devices $\bm{\Omega}$, and phase bias due to manufacturing error $\bm{\Phi}_b \sim \mathcal{U}(0,2\pi)$.

\paragraph{Limited Phase-tuning Control Resolution.} Given the control resolution limits, we can only achieve discretized MZI phase tuning. We assume the phases $\phi$ is uniformly quantized into 8-bit within $[0, 2\pi]$ for phases in $\bm{U}(\bm{\Phi}^U)$, $\bm{\Sigma}(\bm{\Phi}^S)$, $\bm{V}^*(\bm{\Phi}^V)$. 

\paragraph{Phase-shifter Variation.}
We assume the real phase shift $\tilde{\phi}=\frac{\gamma + \Delta \gamma}{\gamma}\phi$, which is proportional to the device-related parameter. We assume $\Delta \gamma \sim \mathcal{N}(0,0.002^2)$. We formulate this error as a diagonal matrix $\boldsymbol{\Gamma}$ multiplied on the phase shift $\boldsymbol{\Phi}'=\boldsymbol{\Gamma} \boldsymbol{\Phi}$.

\paragraph{MZI Crosstalk.}
The crosstalk effect can be modeled as coupling matrix $\boldsymbol{\Omega}$,

\begin{equation}
\begin{aligned}
&\left(\begin{array}{c}
\phi_0^c \\
\phi_1^c \\
\vdots \\
\phi_{N-1}^c
\end{array}\right)=\left(\begin{array}{cccc}
\omega_{0,0} & \omega_{0,1} & \cdots & \omega_{0, N-1} \\
\omega_{1,0} & \omega_{1,1} & \cdots & \omega_{1, N-1} \\
\vdots & \vdots & \ddots & \vdots \\
\omega_{N-1,0} & \omega_{N-1,1} & \cdots & \omega_{N-1, N-1}
\end{array}\right)\left(\begin{array}{c}
\phi_0^v \\
\phi_1^v \\
\vdots \\
\phi_{N-1}^v
\end{array}\right) \\
& \text { s.t. } \omega_{i, j}=1, \quad \forall i=j \\
& \omega_{i, j}=0, \quad \forall i \neq j \text { and } \phi_j \in \mathcal{P} \\
& 0 \leq \omega_{i, j}<1, \quad \forall i \neq j \text { and } \phi_j \in \mathcal{A} .
\end{aligned}
\end{equation}

The diagonal factor $\omega_{i,j}$, $i=j$ is the self-coupling coefficient, $\omega_{i,j}$, $i\neq j$ is the mutual coupling coefficient. We follow~\cite{gu2021l2ight} to assume the self-coupling coefficient to be 1, and the mutual coupling coefficient is 0.005 for adjacent MZIs.

\subsection{Extended Experiment Results} \label{appendix:Detailed Results of Phase-domain Training}
{
 In this section, we provide the extended results of Table \ref{tab:photonics training} and Figure \ref{fig:phase domain}. Each relative $\ell_2$ error takes the form $\text{mean} \pm \text{std}$, where $\text{mean}$ denotes the averaged result over three independent experiments, and $\text{std}$ denotes the corresponding standard deviation. The curves denote averaged relative $\ell_2$ error over three independent experiments and shades denote the corresponding standard deviations.
}

\begin{table}[H]
  \centering
  \caption{Comparison between different photonic training methods. We report the averaged relative $\ell_2$ error and standard deviations across three runs.}
  \resizebox{\linewidth}{!}{
    \begin{tabular}{c|ccc|ccc}
    \toprule
    \toprule
    \multicolumn{1}{r}{} &       & \multicolumn{1}{l}{Black Scholes} &       &       & \multicolumn{1}{l}{20-dim HJB} &  \\
    \cmidrule{2-7}\multicolumn{1}{c}{} & \# MZIs & \# Trainable MZIs & rel. $\ell_2$ error & \# MZIs & \# Trainable MZIs & rel. $\ell_2$ error \\
    \midrule
    \texttt{FLOPS}~\cite{GuDAC} & 18,065 & 18,065 & 0.663$\pm$0.045 & 279,232 & 279,232 & (1.38$\pm$0.07)E-02 \\
    \texttt{L\textsuperscript{2}ight}~\cite{gu2021l2ight} & 18,065 & 2,561 & 0.192$\pm$0.381 & 279,232 & 35,841 & (2.95$\pm$0.99)E-03 \\
    Ours  & \textbf{1,685} & 1,685 & \boldmath{}\textbf{0.114$\pm$0.095}\unboldmath{} & \textbf{2,057} & 2,057 & \boldmath{}\textbf{(2.10$\pm$0.55)E-03}\unboldmath{} \\
    \bottomrule
    \bottomrule
    \end{tabular}%
    
  \label{tab:addlabel}%
  }
\end{table}%

\begin{table}[htbp]
  \centering
  \caption{Comparison between different photonic training methods. We report the averaged relative $\ell_2$ error and standard deviations across three runs.}
    \begin{tabular}{c|ccc|ccc}
    \toprule
    \toprule
    \multicolumn{1}{r}{} &       & \multicolumn{1}{l}{Burgers} &       &       & \multicolumn{1}{l}{Darcy Flow} &  \\
\cmidrule{2-7}    \multicolumn{1}{c}{} & \# MZIs & \# Trainable MZIs & rel. $\ell_2$ error & \# MZIs & \# Trainable MZIs & rel. $\ell_2$ error \\
    \midrule
    \texttt{FLOPS}~\cite{GuDAC} & 72,889 & 72,889 & 0.447$\pm$0.003 & 72,889 & 72,889 & 0.476$\pm$0.002 \\
    \texttt{L\textsuperscript{2}ight}~\cite{gu2021l2ight} & 4,665 & 4,665 & 0.569$\pm$0.003 & 4,665 & 4,665 & 0.154$\pm$0.017 \\
    Ours  & \textbf{2,516} & 2,516 & \boldmath{}\textbf{0.268$\pm$0.010}\unboldmath{} & \textbf{2,516} & 2,516 & \boldmath{}\textbf{0.091$\pm$0.005}\unboldmath{} \\
    \bottomrule
    \bottomrule
    \end{tabular}%
  \label{tab:addlabel}%
\end{table}%

\begin{figure}[H]
\centering
\begin{minipage}[t]{0.45\textwidth}
    \centering
    \subfigure{
        \includegraphics[width=\textwidth]{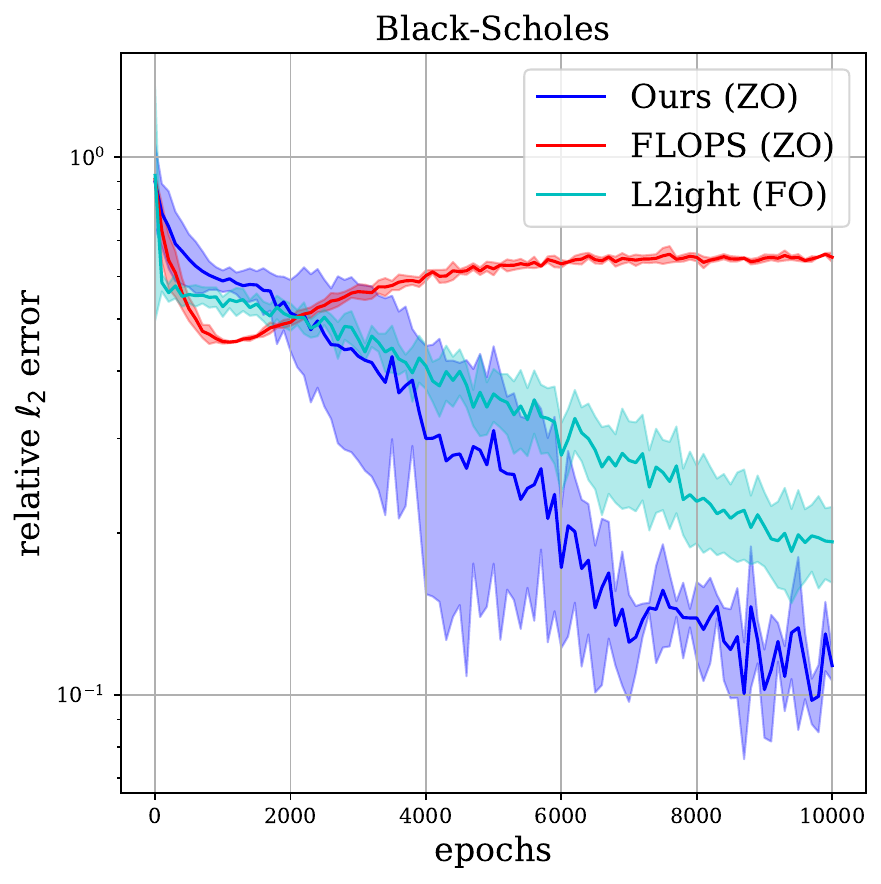}
        \label{fig:apdx_phase_BS}
    }
\end{minipage}
\begin{minipage}[t]{0.45\textwidth}
    \centering
    \subfigure{
        \includegraphics[width=\textwidth]{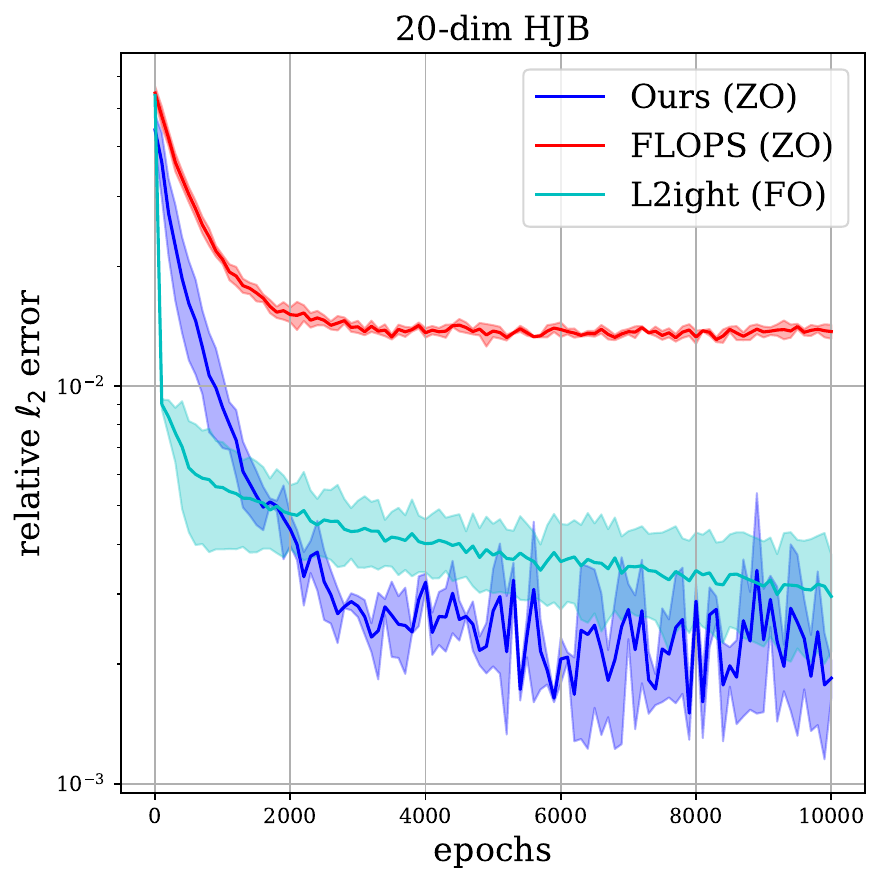}
        \label{fig:apdx_phase_HJB}
    }
\end{minipage}
\quad
\caption{Relative $\ell_2$ error curves of phase domain training for Black-Scholes equation (left) and 20-dim HJB equation (right), respectively. The value at each step is averaged across three runs, and the shade indicates the standard deviation. 
}
\label{fig:abla_loss}
\end{figure}

\begin{figure}[H]
    \centering
    \includegraphics[width=0.8\linewidth]{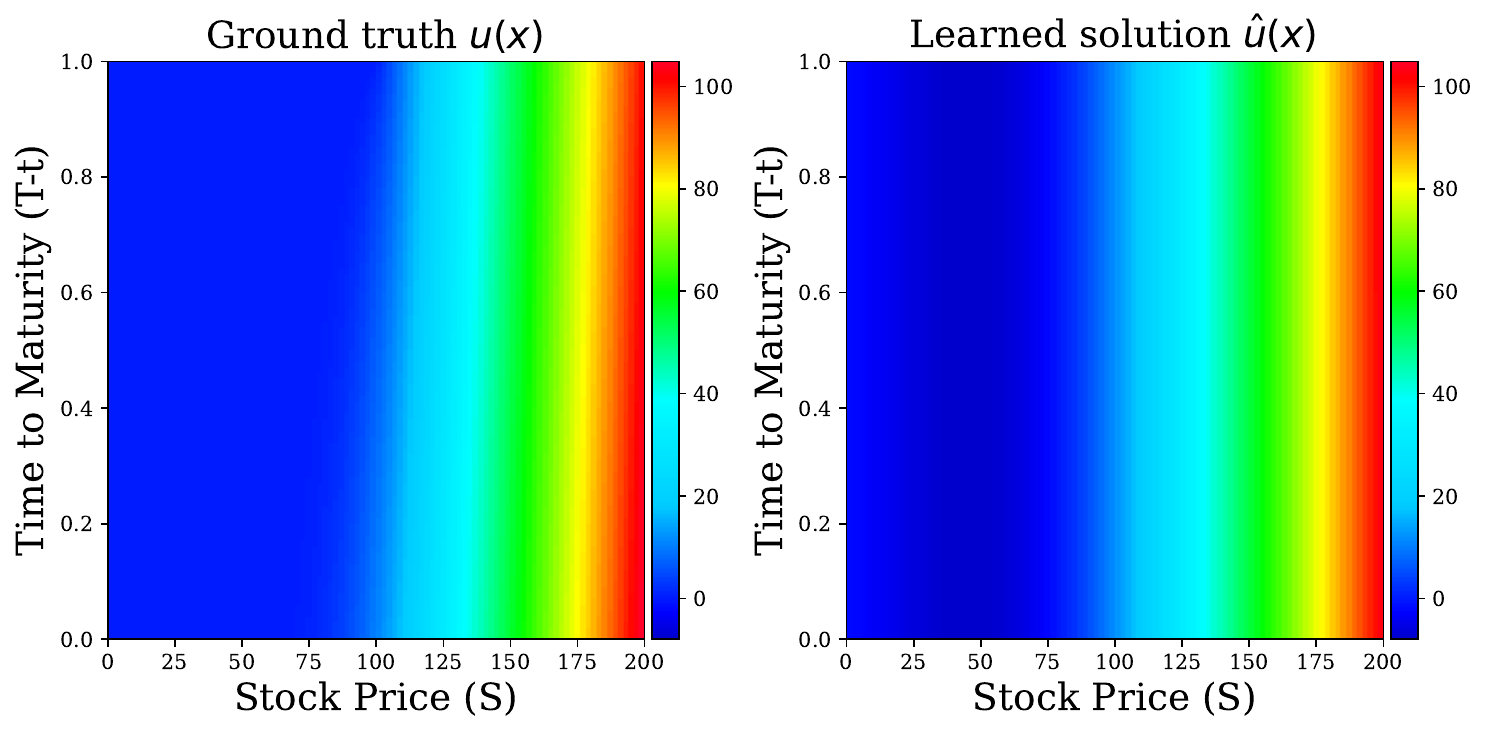}
    \caption{Visualization of Black-Scholes equation in photonic on-chip learning simulation. The left subfigure shows the ground truth $u(x)$, and the right subfigure shows the learned solution $\hat{u}(x)$ using our proposed BP-free PINNs training method.}
    \label{fig:enter-label}
\end{figure}

\subsection{Design Parameters for System Performance Evaluation}\label{appendix:Design Parameters for System Performance Evaluation}

\begin{table}[!h]
\centering
\caption{Design Parameters for System Performance Evaluation.}
\begin{tabular}{ll}
\hline
\textbf{Parameter} & \textbf{Value} \\
\hline
Number of wavelengths & 8 \\
Bit precision ($b$) & 8 \\
MOSCAP MZI length ($L_{\text{MZI}}$) & 1.2 mm \\
8$\times$8 MZI mesh area ($A_{\text{MZI\_mesh}}$) & 16.32 mm$^2$ \\
Comb laser footprint ($A_{\text{Laser}}$) & 0.2 mm$^2$ \\
Microring modulator area ($A_{\text{Modulator}}$) & 0.5 mm$^2$ \\
Photodetector area ($A_{\text{PD}}$) & 0.5 mm$^2$ \\
Cross-connect length ($L_{\text{cross-connect}}$) & 0.2 mm \\
Cross-connect area ($A_{\text{Cross\_connect}}$) & 1.6 mm$^2$ \\
ADC/DAC latency ($t_{\text{ADC}}, t_{\text{DAC}}$) & 24 ns \\
MOSCAP phase shifter tuning delay ($t_{\text{tuning}}$) & 0.1 ns \\
Digital computation ($t_{\text{DIG}}$) & 500 ns \\
\hline
\end{tabular}
\label{tab:apdx-design parameters}
\end{table}

\begin{table}[!h]
  \centering
  \caption{Design parameters for different ONN/TONN layouts.}
    \begin{tabular}{cccc}
    \toprule
    \toprule
          & N     & $n_{MZI_mesh}$ & $n_{cross-connect}$ \\
    \midrule
    ONN-SM & 128   & 256   & / \\
    TONN-SM & 8     & 6     & 1 \\
    \midrule
    ONN-TM & 8     & 1     & / \\
    TONN-TM & 8     & 1     & / \\
    \bottomrule
    \bottomrule
    \end{tabular}%
  \label{tab:addlabel}%
\end{table}%

\section{Broader Impacts} \label{apdx:Broader Impacts}

Real-time PDE solvers on edge devices are desired by many civil and defense applications. However, electronic computing devices fail to meet the requirements.
Our main motivation is to propose a completely back-propagation-free training for PINNs and realize real-time PINNs training (i.e., real-time PDE solver) on photonics computing chips. However, our tensor-train compressed zeroth-order training method can be generally applied to other applications, and our BP-free training framework can be applied to other resource-constrained edge platforms.

\subsection{Extension to Image Classification}
{
 Our tensor-compressed zeroth-order training is a general back-propagation-free training method that applies to lightweight neural networks other than PINNs. 
In this section, we extended our tensor-compressed zeroth-order training to the image classification task on the MNIST dataset. Note that our proposed sparse-grid loss evaluation is designed for PINN training only, so sparse-grid is not used here.

Our baseline model is a two-layer MLP (784$\times$1024, 1024$\times$10) with 814,090 parameters. The dimension of our tensor-compressed training is reduced to 3,962 by folding the input and output layer as size $7\times 4\times 4\times 7\times 8\times 4\times 4\times 8$ and $8\times 4\times 4\times 8\times 1\times 5\times 2\times 1$, respectively. Both the input layer and the output layer are decomposed with a TT-rank $(1, 6, 6, 6, 1)$. Models are trained for 15,000 iterations with a batch size 2,000, using Adam optimizer with an initial learning rate 1e-3 and decayed by 0.8 every 3,000 iterations. In ZO training, we set query number $N=10$ and smoothing factor $\mu=0.01$.

Table \ref{tab:mnist weight} compares results of weight domain training.
\begin{itemize}
    \item Our tensor-train (TT) compressed training does not harm the model expressivity, as TT training achieved a similar test accuracy as standard training in first-order (FO) training.
    \item Our TT compressed training greatly improves the convergence of ZO training and reduces the performance gap between ZO and FO.
\end{itemize}

Table \ref{tab:mnist phase} compares results of phase domain training. Our method outperforms the baseline ZO training method \texttt{FLOPS} \cite{GuDAC}. This is attributed to the tensor-train (TT) dimension reduction that reduced gradient variance. Note that the performance gap between phase domain training and weight domain training could be attributed to the low-precision quantization, hardware imperfections, etc., as illustrated in Section 5.2. 
Our ZO training method did not surpass the FO subspace training method \texttt{L\textsuperscript{2}ight} \cite{gu2021l2ight}.
The performance of \texttt{L\textsuperscript{2}ight} versus our method should be considered case by case. \texttt{L\textsuperscript{2}ight} does not have additional gradient errors due to its FO optimization. Meanwhile, its sub-space training can prevent the solver from achieving a good optimal solution. The real performance depends on the trade-off of these two facts. In our PINN experiments, \texttt{L\textsuperscript{2}ight} underperforms our method because the limitation of its sub-space training plays a dominant role. \texttt{L\textsuperscript{2}ight} performs better on the MNIST dataset, probably because the model is more over-parameterized that even subspace training can achieve a good optimal solution.

The results on the MNIST dataset are consistent with our claims in the submission and support our claim that our method can be extended to image problems with higher dimensions.
}

\begin{table}[H]
  \centering
  \caption{Validation accuracy of weight domain training on MNIST dataset. We report the averaged accuracy and the standard deviation across three runs.}
    \begin{tabular}{ccccc}
    \toprule
    \toprule
    Method & Standard, FO & TT, FO & Standard, ZO & TT, ZO (ours) \\
    \midrule
    Val. Accuracy (\%) & 97.83$\pm$1.02 & 97.26$\pm$0.15 & 83.83$\pm$0.44 & 93.21$\pm$0.46 \\
    \bottomrule
    \bottomrule
    \end{tabular}%
  \label{tab:mnist weight}%
\end{table}%

\begin{table}[H]
  \centering
  \caption{Validation accuracy of phase domain training on MNIST dataset. We report the averaged accuracy and the standard deviation across three runs.}
    \begin{tabular}{cccc}
    \toprule
    \toprule
    \multicolumn{1}{l}{Method} & \texttt{FLOPS} \cite{GuDAC} & \texttt{L\textsuperscript{2}ight} \cite{gu2021l2ight} & ours \\
    \midrule
    Val. Accuracy (\%) & 41.72$\pm$5.50 & 95.80$\pm$0.48 & 87.91$\pm$0.59 \\
    \bottomrule
    \bottomrule
    \end{tabular}%
    
  \label{tab:mnist phase}%
\end{table}%

\subsection{Implementation on other Edge Devices}
Our framework can be easily extended to various resource-constrained hardware platforms. As depicted in Fig. \ref{fig:TONN_training}, the TONN inference accelerator can be replaced by any existing inference accelerator on the edge device. Implementing back-propagation-free training requires only minimal additional modules. These modules can be efficiently implemented either through an external software-based control system (MCU/edge CPU) or directly integrated into the hardware architecture, significantly simpler to design compared to implementing back-propagation computation graphs. BP-free training has been extended for memory-efficient on-device training on microcontrollers \cite{zhao2024poor}, low-cost edge processors (\textit{e.g.,} Raspberry Pi Zero 2) \cite{sugiura2025elasticzo}, smartphones \cite{peng2024pocketllm}, edge GPUs \cite{gao2024enabling}, etc.
We believe our method can also be extended to other resource-constrained hardware platforms including field-programmable gate arrays (FPGAs), application-specific integrated circuits (ASICs), and emerging computing paradigms such as probabilistic circuits. 

\end{document}